\documentclass[a4paper,11pt]{article}
\usepackage{jheppub} 
\usepackage{lineno}
\usepackage{graphicx} 
\usepackage{amsmath}
\usepackage{amsfonts}
\usepackage{cleveref}
\usepackage{booktabs}
\usepackage{xcolor}
\usepackage{subcaption}
\usepackage[ruled,vlined]{algorithm2e}
\usepackage{color}
\usepackage{colortbl}
\usepackage{hyperref}
\usepackage{tikz}
\usepackage{pgffor}
\usepackage{subcaption}
\usepackage{makecell}

\usepackage{amssymb, physics, multirow, array}
\definecolor{lime}{HTML}{A6CE39}
\DeclareRobustCommand{\orcidicon}{\hspace{-2.1mm}
\begin{tikzpicture}
\draw[lime, fill=lime] (0,0) circle [radius=0.13] node[white] {{\fontfamily{qag}\selectfont \tiny \,ID}}; \draw[white, fill=white] (-0.0525,0.095) circle [radius=0.007]; 
\end{tikzpicture} \hspace{-3.2mm} }
\foreach \x in {A, ..., Z}{\expandafter\xdef\csname orcid\x\endcsname{\noexpand\href{https://orcid.org/\csname orcidauthor\x\endcsname} {\noexpand\orcidicon}}}

\title{\boldmath Amortized Inference of Multi-Modal Posteriors using Likelihood-Weighted Normalizing Flows}

\author{Rajneil Baruah\orcidA{}}
\affiliation{Department of Physics, SEAS, Bennett University, Greater Noida, Uttar Pradesh, India, 201310}
\affiliation{Department of Physics, Bangabasi Evening College, Kolkata, West Bengal, 700011}

\emailAdd{rajneilb.physics@gmail.com}

\abstract{We present a technique for posterior estimation using Normalizing Flows trained with likelihood-weighted importance sampling, the special case of neural importance sampling in which the proposal is the prior. This approach allows for the efficient inference of theoretical parameters in high-dimensional inverse problems without the need for posterior training samples. We implement the method on multi-modal benchmark tasks in 2D and 3D to check for the efficacy. A critical observation of our study is the impact of the base distribution on the modelled posteriors. We find that standard unimodal base distributions leave spurious probability \textit{bridges} between well-separated modes. We demonstrate that initializing the flow with a Gaussian Mixture Model that matches the number of target modes significantly improves reconstruction fidelity, as measured by some distance and divergence metrics. Finally, we apply this method to a curated problem in heavy flavour physics --- the extraction of the Wolfenstein parameters from the CP asymmetry in $B^0\to J/\psi\,K^0$; it is multimodal,
non-Gaussian, and asymmetric in its mode weights --- and compare the results against a well-converged Markov Chain Monte Carlo reference and an exact reference computed by quadrature, using different
metrics. Used as a proposal for importance sampling, the trained flow removes its remaining bias and gives the evidence.}

\begin{document}

\maketitle
\flushbottom

\section{Introduction}
\label{sec:intro}
Across diverse domains—from complex systems and finance to high-energy physics and astrophysics—scientific inquiry often relies on deriving theoretical parameters from observational data~\cite{tarantola2005}. At the core of this challenge lies the \textit{inverse problem}: inferring the posterior distribution of theoretical parameters given a set of observables~\cite{cranmer2020}.

Traditional approaches for posterior estimation rely on sampling algorithms such as Markov Chain Monte Carlo (MCMC)~\cite{metropolis1953, hastings1970} and Nested Sampling (NS)~\cite{skilling2006}. In astrophysics and cosmology, implementations like \texttt{emcee}~\cite{foreman2013} and \texttt{dynesty}~\cite{speagle2020} have become standard tools. While these frameworks are statistically robust, they suffer significantly from the \textit{curse of dimensionality}. In real-world scenarios, where the parameter space is high-dimensional and the likelihood function relies on computationally expensive simulators (e.g., in particle physics phenomenology~\cite{brehmer2018}), convergence can take weeks or even months.

Recent advances in machine learning have introduced \textit{Normalizing Flows (NFs)} as a powerful alternative for probabilistic modelling~\cite{tabak2010, rezende2015}. By learning a bijective mapping between a simple base distribution (e.g., a Gaussian) and the complex target distribution, NFs allow for exact density estimation and efficient sampling~\cite{papamakarios2021} from the target distribution. Modern architectures, such as RealNVP~\cite{dinh2016} and Neural Spline Flows~\cite{durkan2019}, offer enough expressivity to model highly complex distributions.

However, standard training methodologies for NFs (typically Maximum Likelihood Estimation) rely on a critical assumption: that we possess a dataset of samples drawn \textit{from the target posterior}~\cite{papamakarios2017}. In the context of Simulation-Based Inference (SBI)~\cite{cranmer2020, lueckmann2021}, this often leads to approaches like Neural Posterior Estimation (NPE)~\cite{papamakarios2016, greenberg2019}, which still require substantial simulation budgets to generate training data.

In many scientific inverse problems, this assumption does not hold. We generally possess only a generic understanding of the parameters, hence assume they belong to a uniform distribution and a ``black-box'' simulator that can evaluate the likelihood of a given point. If we simply train a Normalizing Flow on samples drawn from the prior, the network will trivially learn to reconstruct the prior, failing to capture the information contained in the likelihood. While Variational Inference (VI) techniques exist to approximate posteriors~\cite{blei2017}, applying them efficiently with flows to capture multi-modal structures remains an active area of research~\cite{wirnsberger2020, midgley2020}. Other recent efforts for estimating posterior distributions with machine learning efforts can be can be seen in~\cite{Baruah:2024gwy, Baruah:2025nby}

In this work, we study a method to bridge this gap. By leveraging likelihood evaluations as \textit{sample weights} during training—akin to Neural Importance Sampling~\cite{muller2019}—we demonstrate that a Normalizing Flow can directly learn the transformation from a simple base density to the complex target posterior, without ever seeing a ``true'' posterior sample. The objective itself is not new: it is that of neural importance sampling with the proposal fixed at the prior (\cref{sec:nis}). Our contribution is the study of what this simplest form of the method can and cannot do, on benchmarks with known answers and on a physics problem with an exact answer, and the use of the trained flow as a proposal that corrects its own errors.

\section{Theoretical Framework: Likelihood-Weighted Normalizing Flows}
\label{sec:Theory}

\subsection{Normalizing Flows}
Let $f_\phi: \mathcal{Z} \to \Theta$ be a differentiable bijection with a differentiable inverse (i.e., a diffeomorphism), parameterized by neural network weights $\phi$. We consider a latent space $\mathcal{Z} \subseteq \mathbb{R}^D$ endowed with a known, tractable base density $p_{\text{\small Z}}(z)$ (e.g., a standard Normal distribution or a Mixture of Gaussians), and a target parameter space $\Theta \subseteq \mathbb{R}^D$ associated with the unknown target distribution $p(\theta)$ (or the posterior $p(\theta|x)$)~\cite{tabak2010, rezende2015, papamakarios2021}.

The normalizing flow $f_\phi$ transforms the probability mass from the base distribution to the target. Because $f_\phi$ is a diffeomorphism, it maps the support of the base onto the support of $q_\phi$ without changing its topology. For a Gaussian base and a Gaussian-mixture target, both supports are the whole of $\mathbb{R}^D$, so this is not a strict obstruction. However, separating well-separated modes requires the flow to stretch a small region of the base over the low-density region between them, and a flow can approximate a target whose support differs topologically from that of its base only if its bi-Lipschitz constant grows without bound~\cite{cornish2020}. This limitation, rather than a strict topological constraint, is what we observe in \cref{sec:implementationandresults}. 

The density of the transformed variable $\theta = f_\phi(z)$ is obtained analytically via the change of variables formula:
\begin{align}
\label{eq:q_def1}
q_\phi(\theta) = p_{\text{\small Z}}(z)\left(z\right)\left|\det\!\left(\frac{\partial z}{\partial \theta}\right)\right|
\end{align}
Since $f_{\phi}(z)$ is bijective, we can substitute $z=f_\phi^{-1}(\theta)$:
\begin{align}
q_\phi(\theta)
&= p_{\text{\small Z}}(z)\big(f_\phi^{-1}(\theta)\big)\,\left|\det\!\left(\frac{\partial f_\phi^{-1}(\theta)}{\partial \theta}\right)\right|
\label{eq:q_def2}
\end{align}
Expressing this in logarithmic form yields:
\begin{align}
\label{eq:logq_jac}
\log q_\phi(\theta)
&= \log p_{\text{\small Z}}(z)\big(f_\phi^{-1}(\theta)\big) + \log\left|\det\!\left(\frac{\partial f_\phi^{-1}(\theta)}{\partial \theta}\right)\right|
\end{align}

To optimize the parameters $\phi$, we look to the principle of Maximum Likelihood Estimation (MLE). We seek to maximize the likelihood of the model generating the observed data, which is equivalent to minimizing the Kullback-Leibler (KL) divergence between the true distribution and the model approximation.

\subsection{KL Divergence, Maximum Likelihood Estimation and the Loss Function}
The KL divergence from a reference probability distribution \(r(\theta)\) to a model distribution \( q_\phi(\theta)\) is defined as~\cite{1320776d-9e76-337e-a755-73010b6e4b64}:
\begin{align}
\label{KLDiv}
    D_{\text{\small KL}}\left(r(\theta)||q_\phi(\theta)\right) &= \int r(\theta) \log\frac{r(\theta)}{q_\phi(\theta)} d\theta \nonumber \\
    &= \int r(\theta)\log(r(\theta))d\theta - \int r(\theta) \log(q_\phi(\theta)) d\theta \nonumber \\
    &= \mathbb{E}_{\theta \sim r(\theta)} \left[ \log(r(\theta)) \right] -\mathbb{E}_{\theta \sim r(\theta)} \left[ \log(q_\phi(\theta)) \right].
\end{align}
If we consider training samples \(\{\theta_i\}_{i=1}^N\) drawn i.i.d.\ from the distribution \(r(\theta)\), the empirical negative log-likelihood (NLL) is given by:
\begin{align}
\label{negLLdef}
\mathcal{L}(\phi)
&= -\frac{1}{N}\sum_{i=1}^N \log q_\phi(\theta_i)
\quad\stackrel{N\to\infty}{\longrightarrow}\quad
- \mathbb{E}_{\theta\sim r(\theta)}\left[ \log q_\phi(\theta) \right].
\end{align}
Comparing \cref{negLLdef} and \cref{KLDiv}, we observe that:
\begin{align}
    \mathcal{L}(\phi) = D_{\text{\small KL}}\left(r(\theta)\Vert q_{\phi} (\theta)\right) - \mathbb{E}_{\theta \sim r(\theta)} \left[ \log(r(\theta)) \right].
\end{align}
Since the entropy term $\mathbb{E}\left[ \log(r(\theta)) \right]$ is independent of $\phi$, minimizing the negative log-likelihood is mathematically equivalent to minimizing the forward KL divergence ~\cite{papamakarios2021}:
\begin{align}
\label{KLequivLoss}
   \min_\phi \left( \mathcal{L}(\phi) \right) \iff \min_\phi \left( D_{\text{\small KL}}\left(r(\theta)\Vert q_{\phi} (\theta)\right) \right).
\end{align}

\subsection{Application to Posterior Estimation}
In the context of Bayesian inference, our goal is to model the posterior distribution $p(\theta | D)$. According to Bayes' Theorem:
\begin{align}
p(\theta \mid D) = \frac{p(D \mid \theta) \, p(\theta)}{p(D)} \,,
\end{align}
where \(p(D \mid \theta)\) is the likelihood, \(p(\theta)\) is the prior, and \(p(D) = Z\) is the evidence. Substituting $r(\theta) = p(\theta|D)$ into our objective function:
\begin{align}\label{eq:minimization1}
    D_{\text{\small KL}}\left( p(\theta \mid D)\Vert q_{\phi} (\theta)\right) &= \int  p(\theta \mid D) \log\frac{p(\theta \mid D)}{q_{\phi} (\theta)} d\theta \nonumber \\
    &= \int  \frac{1}{Z}p(\theta) p(D \mid \theta) \log\frac{p(\theta \mid D)}{q_{\phi} (\theta)} d\theta \nonumber \\
    &= \mathbb{E}_{\theta \sim p(\theta)} \left[ \frac{1}{Z} p\left( D \mid \theta \right) \log\left( p\left( \theta \mid D\right) \right) \right] \nonumber \\
    & \qquad -\mathbb{E}_{\theta \sim p(\theta)} \left[ \frac{1}{Z}p\left( D \mid \theta \right) \log\left( q_{\phi}(\theta) \right) \right].
\end{align}
Again, retaining only the terms dependent on $\phi$, minimizing the divergence is equivalent to minimizing the likelihood-weighted negative log-likelihood:
\begin{align}\label{eq:minimization2}
   \min_\phi D_{\text{\small KL}}\left( p(\theta \mid D)\Vert q_{\phi}(\theta)\right) \iff \min_\phi \left( - \mathbb{E}_{\theta \sim p(\theta)} \left[ \frac{p(D \mid \theta)}{Z} \log\left( q_{\phi}(\theta) \right) \right] \right).
\end{align}
In practice, we approximate this expectation using $N$ samples drawn from the prior $p(\theta)$, weighted by their data-likelihood $L(\theta_i) \equiv p(D|\theta_i)$. The final loss function becomes:
\begin{align}
\label{eq:LossFinal}
    \mathcal{L}(\phi) = -\frac{1}{N} \sum_{i=1}^N \left[ L(\theta_i) \log \left( q_{\phi}(\theta_i)\right) \right].
\end{align}

\section{Comparison with Other ML-Based Methods}
\label{sec:comparison}

Several established machine learning approaches exist for posterior
estimation, including variational inference~\cite{blei2017variational,
rezende2015variational}, importance-weighted variational
inference~\cite{burda2016importance}, and simulation-based
inference~\cite{cranmer2020frontier, lueckmann2021benchmarking}. The method studied in this work shares the use of normalizing
flows~\cite{papamakarios2021normalizing} and of likelihood evaluations with
each of these, but its objective is different. It frames posterior
inference as a likelihood-weighted density estimation problem, without
posterior samples, simulator budgets, or bounds on the marginal
likelihood. This objective is that of neural importance
sampling~\cite{muller2019neural} with the proposal fixed at the prior
(\cref{sec:nis}). In this section, we make these relations precise.

\subsection{Consistency of the Likelihood-Weighted Objective}

We first show that the likelihood-weighted objective recovers the true
posterior when the model family is sufficiently expressive.

Let $q_\phi(\theta)$ denote the density represented by the normalizing
flow, and let the posterior be $p(\theta|D) \propto p(D|\theta)p(\theta)$.
The objective of \cref{eq:minimization2} is the \emph{forward} divergence,
\begin{equation}
  D_{KL}\!\left(p(\theta|D)\,\big\|\,q_\phi(\theta)\right)
  = -\frac{1}{Z}\,\mathbb{E}_{p(\theta)}\!\left[p(D|\theta)\log q_\phi(\theta)\right] + \text{const},
  \label{eq:forwardkl_consistency}
\end{equation}
where the constant does not depend on $\phi$. Since
$D_{KL}(p\|q) \ge 0$~\cite{kullback1951information}, with equality if and only
if $q=p$ almost everywhere, the global minimum is reached when
$q_\phi(\theta) = p(\theta|D)$, provided the model class is expressive
enough.

We stress what the objective is not. The \emph{reverse}
divergence,
\begin{equation}
  D_{KL}(q_\phi \| p(\theta|D))
  = \mathbb{E}_{q_\phi}\!\left[
      \log q_\phi(\theta) - \log p(D|\theta) - \log p(\theta)
    \right] + \text{const},
\end{equation}
is the objective of variational inference (\cref{sec:vi}). It is an
expectation under the flow, in which the likelihood enters inside a
logarithm, and it does not reduce to the weighted loss of
\cref{eq:LossFinal}. The two divergences share the same minimum for an
infinitely flexible flow, but they behave differently for a flow of finite
capacity. The reverse divergence is mode-seeking~\cite{blei2017variational}.
The forward divergence is mass-covering: a region where the posterior is
large and the flow is small is penalised strongly, but mass that the flow
places where the posterior is small is penalised only weakly, through the
normalisation of $q_\phi$~\cite{bishop2006}. We return to this tolerance of mass between
the modes in \cref{sec:implementationandresults}.

\subsection{Connection to Importance Sampling}
\label{sec:nis}

The training procedure can be interpreted through importance
sampling~\cite{muller2019neural}. Drawing samples from a proposal $r(\theta)$, expectations
under the posterior can be written as
\begin{equation}
  \mathbb{E}_{p(\theta|D)}[f(\theta)]
  = \frac{\mathbb{E}_{r(\theta)}\!\left[f(\theta)\,w(\theta)\right]}
         {\mathbb{E}_{r(\theta)}\!\left[w(\theta)\right]},
  \qquad
  w(\theta) = \frac{p(D|\theta)\,p(\theta)}{r(\theta)}.
\end{equation}
When the proposal is the prior, $r(\theta)=p(\theta)$, the weights
simplify to $w(\theta)\propto p(D|\theta)$.  The method proposed here
uses these likelihood-weighted prior samples to train the normalizing
flow, which is therefore a form of density estimation under an
importance-reweighted measure.

This places the method in a known family. Weighting prior samples by
their likelihood is \textit{likelihood weighting}~\cite{fung1989}. Fitting
a parametric density to the weighted sample by maximising the weighted
log-likelihood has the form of one iteration of the cross-entropy
method~\cite{rubinstein1999}. With a flow as the density, it is the
objective of NIS~\cite{muller2019neural}, which has been applied to the
integration of phase space in particle physics~\cite{gao2020iflow,
bothmann2020}. In NIS, the proposal is built from the flow and is updated
as the flow improves, with weights $p(D|\theta)p(\theta)/r(\theta)$. The
method we study here is the special case in which the proposal is frozen at
the prior. We measure its efficiency by the effective sample size
$N_{\rm eff} = (\sum_i w_i)^2/\sum_i w_i^2$~\cite{kish1965}. For large $N$
and $r(\theta)=p(\theta)$,
\begin{equation}
  \frac{N_{\rm eff}}{N} \;\to\;
  \frac{\mathbb{E}_{p(\theta)}[w]^2}{\mathbb{E}_{p(\theta)}[w^2]}
  = \frac{1}{1+\chi^2\!\left(p(\theta|D)\,\|\,p(\theta)\right)},
  \label{eq:ess_chi2}
\end{equation}
where $\chi^2(p\|q) = \int (p-q)^2/q\;d\theta$ is the chi-squared
divergence. The fraction of useful prior samples is therefore set by how
concentrated the posterior is relative to the prior, and it can be very
small. On the other hand, a trained flow is a normalised density, so it
can in turn be used as the proposal of an importance-sampling step that
corrects its errors (\cref{sec:bmeson:is}).

\subsection{Relation to Variational and Importance-Weighted Inference}

The proposed method is related to variational inference~(VI)~\cite{blei2017variational},
importance-weighted variational inference~(IWVI)~\cite{burda2016importance}, and
simulation-based inference~(SBI)~\cite{cranmer2020frontier}, but differs from each in the
objective being optimized and in the role played by the likelihood.

\subsubsection{Variational Inference}
\label{sec:vi}

VI~\cite{blei2017variational, rezende2015variational} approximates $p(\theta|D)$ by maximising the Evidence Lower Bound
(ELBO):
\begin{equation}
  \mathcal{L}_{\mathrm{ELBO}}
  = \mathbb{E}_{q_\phi(\theta)}\!\left[
      \log p(D|\theta) + \log p(\theta) - \log q_\phi(\theta)
    \right].
\end{equation}
Here the likelihood $p(D|\theta)$ appears \emph{inside a logarithm}
and the expectation is taken \emph{under the model} $q_\phi$.  The
objective simultaneously fits the likelihood and regularises toward
the prior.

By contrast, the proposed method uses
\begin{equation}
  \max_\phi \;
  \mathbb{E}_{\theta \sim r(\theta)}\!\left[
    p(D|\theta)\,\log q_\phi(\theta)
  \right],
\end{equation}
where $r(\theta)$ is a fixed proposal (e.g.\ the prior) and the
expectation is taken \emph{under $r$}, not under $q_\phi$.  The
likelihood acts as a \emph{weight on samples} rather than entering the
objective logarithmically.  The method is therefore a
\emph{likelihood-weighted density-estimation problem}, not a lower
bound on the marginal likelihood.

\subsubsection{Importance-Weighted Variational Inference}

IWVI~\cite{burda2016importance} tightens the ELBO by using $K$ samples:
\begin{equation}
  \log p(D) \ge
  \mathbb{E}_{\theta_{1:K}\sim q_\phi}\!\left[
    \log \frac{1}{K}\sum_{k=1}^{K}
    \frac{p(D|\theta_k)\,p(\theta_k)}{q_\phi(\theta_k)}
  \right].
\end{equation}
The importance weights $w_k = p(D|\theta_k)p(\theta_k)/q_\phi(\theta_k)$
depend explicitly on the \emph{model density} $q_\phi$, and the
objective is a bound on $\log p(D)$.

The proposed method differs in three respects:
\begin{itemize}
  \item the sample weights do not involve $q_\phi(\theta)$;
  \item no bound on $\log p(D)$ is constructed or optimised;
  \item $q_\phi$ is fitted directly to the likelihood-weighted
        distribution $\tilde{p}(\theta)\propto p(D|\theta)\,r(\theta)$,
        which equals the posterior when $r(\theta)=p(\theta)$.
\end{itemize}
Equivalently, the method minimises the forward divergence
$D_{KL}(\tilde{p}\|q_\phi)$ rather than a marginal-likelihood bound.

\subsubsection{Simulation-Based Inference}

Neural Posterior Estimation~(NPE)~\cite{papamakarios2016fast, greenberg2019automatic},
a leading SBI method~\cite{cranmer2020frontier, lueckmann2021benchmarking},
learns a conditional density $q_\phi(\theta|x)$ by minimising
\begin{equation}
  \mathbb{E}_{p(\theta,x)}\!\left[-\log q_\phi(\theta|x)\right],
\end{equation}
which requires sampling from the joint simulator distribution
$p(\theta,x)$.

The proposed method differs in that:
\begin{itemize}
  \item it does not require samples from $p(\theta|D)$ or $p(\theta,x)$;
  \item it uses direct pointwise evaluations of $p(D|\theta)$;
  \item posterior estimation reduces to a weighted density-estimation
        problem, without any simulation budget for the observations.
\end{itemize}

\subsection{Summary}

Table~\ref{tab:method_comparison} summarises the objective functions of
the five approaches.

\begin{table}[h]
  \centering
  \begin{tabular}{l l}
    \hline
    Method & Objective \\
    \hline
    VI   & $\mathbb{E}_{q}[\log p(D|\theta)]
           - D_{KL}(q\,\|\,p(\theta))$ \\[4pt]
    IWVI & $\mathbb{E}_{q}\!\left[
             \log\frac{1}{K}\sum_k
             \frac{p(D|\theta_k)p(\theta_k)}{q(\theta_k)}
           \right]$ \\[4pt]
    SBI  & $\mathbb{E}_{p(\theta,x)}\!\left[-\log q_\phi(\theta|x)\right]$ \\[4pt]
    NIS  & $\mathbb{E}_{r(\theta)}\!\left[
             \frac{p(D|\theta)\,p(\theta)}{r(\theta)}\,\log q_\phi(\theta)
           \right]$, $r$ built from $q_\phi$ \\[4pt]
    LW-NF & $\mathbb{E}_{p(\theta)}\!\left[
                 p(D|\theta)\,\log q_\phi(\theta)
               \right]$ \\
    \hline
  \end{tabular}
  \caption{Comparison of objective functions across methods. LW-NF
           performs likelihood-weighted density estimation rather than
           evidence maximisation or conditional density learning. It is the
           special case of NIS in which the proposal $r$ is the prior.}
  \label{tab:method_comparison}
\end{table}

The key distinction across these comparisons is that the method
recasts posterior inference as a density-estimation problem, using the
likelihood only as a reweighting factor on prior samples. This sidesteps
the need for posterior samples, simulator budgets, or marginal-likelihood
bounds, which makes the approach attractive in settings where the
likelihood can be evaluated pointwise but posterior samples are
unavailable. The price is that of any importance sampler with a fixed
proposal, \cref{eq:ess_chi2}: the training set is only as informative as
its effective sample size.

\section{Implementation and Results}
\label{sec:implementationandresults}

To evaluate the efficacy of the proposed likelihood-weighted training, we apply the method to a set of synthetic toy problems in 2 and 3 dimensions. These problems are designed with varying degrees of topological complexity (in terms of the number of modes) to test the expressivity of the flow. To quantify the fidelity of the posterior reconstruction, we report the Kullback-Leibler ($D_{\text{\small KL}}$) Divergence and the Average Marginal Wasserstein Distance between the true and modelled distributions.

\subsection{Methodology}
\label{sec:methodology}

We define a $d$-dimensional parameter space $\theta \in \mathbb{R}^d$ governed by a prior $\pi(\theta)$ (typically uniform) and a likelihood function $L(\theta) \equiv p(D|\theta)$ given observed data $D$.

\begin{figure}
    \centering
    \includegraphics[width=\textwidth]{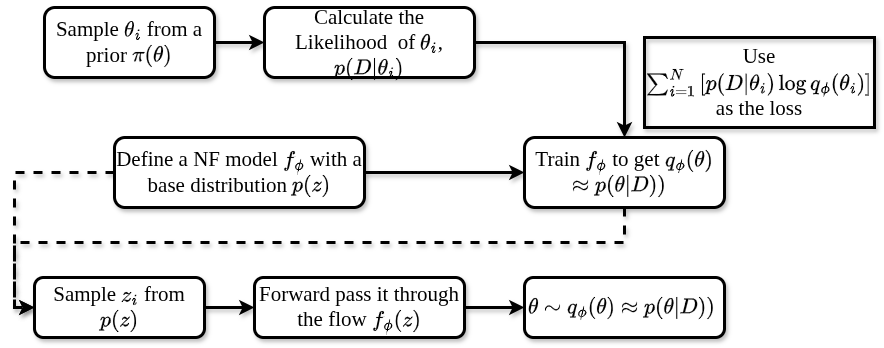}
    \caption{\small The workflow, including training the NF and then sampling from the base to pass it through the flow}
    \label{fig:flowchart}
\end{figure}

The training procedure is outlined in Algorithm \ref{alg:training}. We first generate a static dataset of $N$ samples drawn from the prior, $\{\theta_i\}_{i=1}^N \sim \pi(\theta)$, and evaluate their corresponding likelihoods $w_i = L(\theta_i)$. These likelihoods serve as importance weights. We then train a Normalizing Flow $f_\phi: \mathcal{Z} \to \Theta$ to map a base latent distribution $p_{\text{\small Z}}(z)$ to the target posterior. The network is optimized by minimizing the likelihood-weighted negative log-likelihood (see \cref{eq:LossFinal}). The workflow and the algorithm are shown in \cref{fig:flowchart}.

\begin{algorithm}[H]
\SetAlgoLined
\KwResult{Trained Flow parameters $\phi^*$ approximating posterior $p(\theta|D)$}
 \textbf{Inputs:} Prior $\pi(\theta)$, Likelihood $L(\theta)$, Base distribution $p_Z(z)$\;
 \textbf{Data Generation:}\\
 Sample $\Theta_{\text{train}} = \{\theta_1, \dots, \theta_N\} \sim \pi(\theta)$\;
 Compute Weights $W = \{w_1, \dots, w_N\}$ where $w_i = L(\theta_i)$\;
 \textit{Optional:} Pre-process $W$ (e.g., clipping) for numerical stability\;
 \textbf{Training:}\\
 \While{not converged}{
  Sample batch $(\theta_b, w_b)$ from $(\Theta_{\text{train}}, W)$\;
  Compute $z_b = f_\phi^{-1}(\theta_b)$\;
  Compute $\mathcal{L}(\phi) = - \frac{1}{|b|} \sum w_i \log q_\phi(\theta_i)$\;
  Update $\phi \gets \phi - \eta \nabla_\phi \mathcal{L}$\;
 }
 \caption{Likelihood-Weighted Flow Training}
 \label{alg:training}
\end{algorithm}

\subsection{Implementation Details}
\label{sec:implementation}

We utilize the \texttt{normflows} library~\cite{Stimper2023} based on \texttt{PyTorch}. As the loss function is architecture-agnostic, we employ the Real Non-Volume Preserving (RealNVP) architecture for all experiments. Unless otherwise stated, the base distribution $p_Z(z)$ is a standard multivariate Gaussian.

\subsubsection{2-Dimensional Benchmarks}
\begin{figure}[htbp]
\centering
\includegraphics[width=.99\textwidth, height = 5cm]{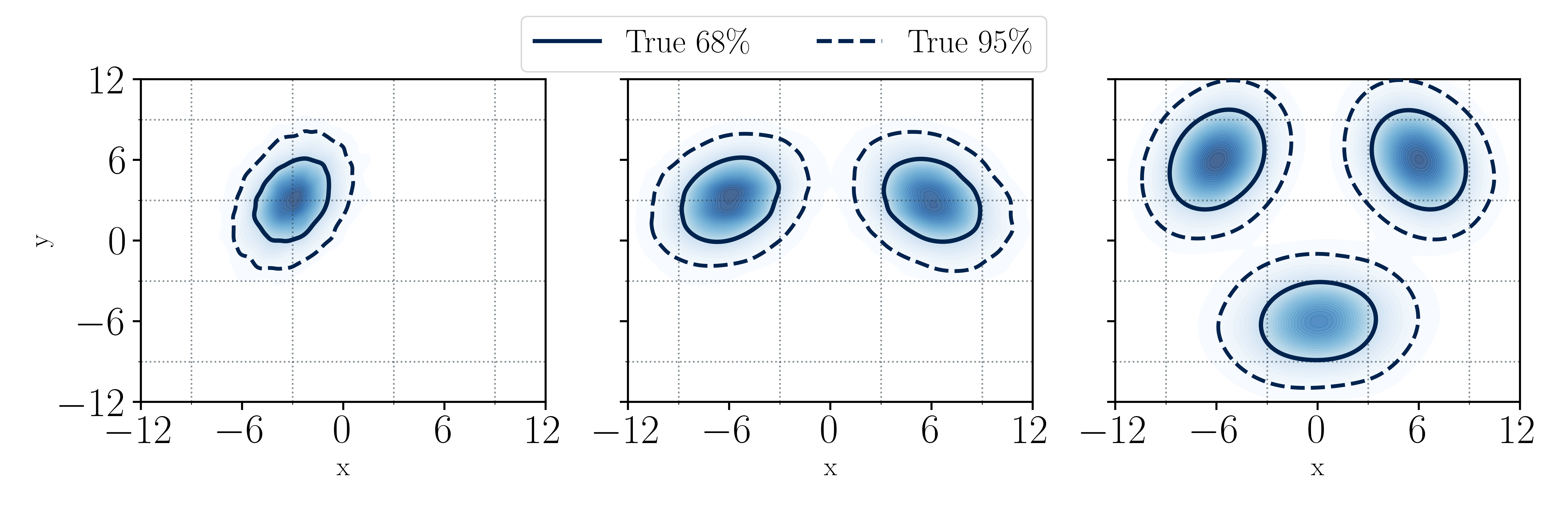}
\caption{\footnotesize Ground truth densities for the 2D benchmark tasks. From left to right: Single-mode, Two-mode, and Three-mode posteriors constructed from Gaussian mixtures.}
\label{fig:2DPrior}
\end{figure}

We define a uniform prior over the domain $\Omega = [-12, 12]^2$. To rigorously test mode coverage, we construct three synthetic posteriors with increasing multimodality (see \cref{fig:2DPrior}):

\begin{enumerate}
    \item \textbf{Single-mode:} A single Gaussian centred at $\boldsymbol{\mu}_1 = [-3, 3]^\top$.
    \item \textbf{Two-mode:} An equal mixture of two Gaussians centred at $\boldsymbol{\mu}_{2,1} = [-6, 3]^\top$ and $\boldsymbol{\mu}_{2,2} = [6, 3]^\top$.
    \item \textbf{Three-mode:} An equal mixture of three Gaussians located at $[-6, 6]^\top$, $[6, 6]^\top$, and $[0, -6]^\top$.
\end{enumerate}
The covariance matrices are chosen to introduce correlation between dimensions, ensuring the task is not trivially separable.

\begin{figure}[htbp]
\centering
\includegraphics[width=.99\textwidth, height = 5cm]{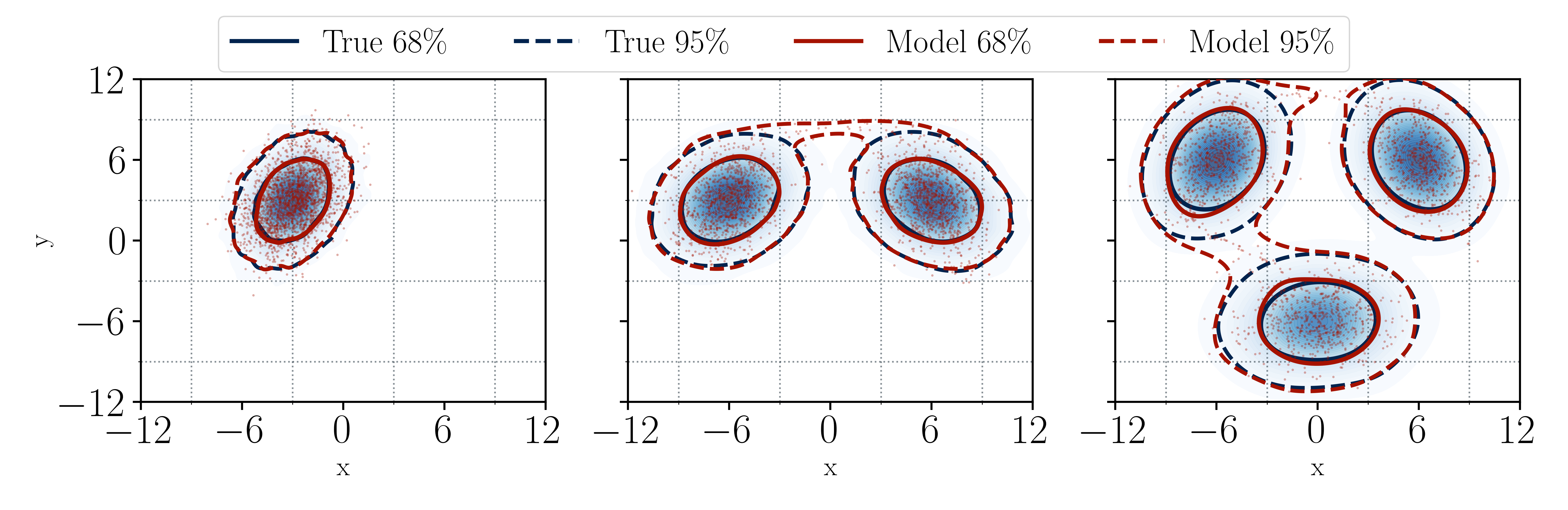}
\caption{\footnotesize Modelled posterior distributions $q_\phi(\theta)$ generated by the Normalizing Flow. While the single-mode posterior (Left) is captured accurately, the multi-modal cases (Center, Right) exhibit spurious ``bridges'' connecting the modes.}
\label{2DtrainedPlots}
\end{figure}

For each scenario, we train a RealNVP model with a standard Gaussian base distribution. \Cref{2DtrainedPlots} illustrates the density of samples generated by the trained flows.

\paragraph{Quantitative Analysis}
We report the KL Divergence (approximated via Monte Carlo integration) and the Average Marginal Wasserstein-1 Distance ($\overline W_{1}$) in \cref{tab:metrics_2D_singlebase}.

\begin{table}[h!]
\centering
\small
\begin{tabular}{lccc}
\toprule
\textbf{Target Posterior} & $D_{\text{\small KL}}$ & $\overline W_{1}$
 & $\overline W_{1}$ (floor) \\
\midrule
Single-Mode & $<10^{-3}$ & 0.0287 & 0.0486 \\
Two-Mode    & 0.0143     & 0.1154 & 0.1009 \\
Three-Mode  & 0.0290     & 0.1352 & 0.0845 \\
\bottomrule
\end{tabular}
\caption{\small Quantitative comparison of posterior reconstruction with the same model (single-Gaussian base) against three target posteriors of increasing multimodality. The \emph{noise floor} is the same $\overline W_1$ computed between two disjoint halves of the reference sample; it is the value that two samples of the \emph{identical} distribution produce at this sample size, and a reported distance is meaningful only insofar as it exceeds it. The single-mode reconstruction is at the floor, i.e.\ statistically indistinguishable from the truth, while the multimodal cases sit clearly above it.}
\label{tab:metrics_2D_singlebase}
\end{table}

The metrics reveal a clear trend: performance degrades as the complexity of the true posterior increases. While the KL divergence remains low ($<0.05$), indicating good global overlap, the Wasserstein distance increases significantly for the multimodal cases. We can see the reason in \cref{2DtrainedPlots}: between the modes, the flow leaves thin bridges of probability, which carry little mass but are far from where the mass should be. These bridges have a simple origin. The supports of the Gaussian base and of the Gaussian-mixture target are both the whole plane, so there is no strict topological obstruction. However, to empty the region between two separated modes, the flow has to stretch a thin slice of the base over that whole region, which requires a very large Jacobian. A flow can approximate a target whose support is topologically different from that of its base only if its bi-Lipschitz constant grows without bound~\cite{cornish2020}, and a target whose modes are separated by regions of exponentially small density approaches this limit. A trained flow of finite depth can only stretch its base by a finite amount, and the residual mass is left as a bridge. The same difficulty of flows with a unimodal base on multimodal targets led Stimper \textit{et al.} to learn the base distribution itself~\cite{stimper2022}. The likelihood-weighted loss does little to suppress the bridges: since it estimates the forward divergence, mass placed between the modes, where the posterior is small, is penalised only weakly (\cref{sec:comparison}). This is why we find a small KL divergence together with a growing Wasserstein distance.

\subsubsection{3-Dimensional Benchmark}
We extend the analysis to $\mathbb{R}^3$ with a three-mode Gaussian mixture posterior defined as:
\begin{align}
\label{eq:3DTruePosterior}
p(\boldsymbol{\theta}) = \sum_{k=1}^{3} \frac{1}{3}\mathcal{N}(\boldsymbol{\theta} | \boldsymbol{\mu}_k, \boldsymbol{\Sigma}_k),
\end{align}
with means located at $[5, 5, -5]^\top$, $[5, 7, 5]^\top$, and $[-5, 7, 5]^\top$, and full rank covariance matrices.

\begin{figure}[htbp]
\centering
\includegraphics[width=.99\textwidth, height = 5cm]{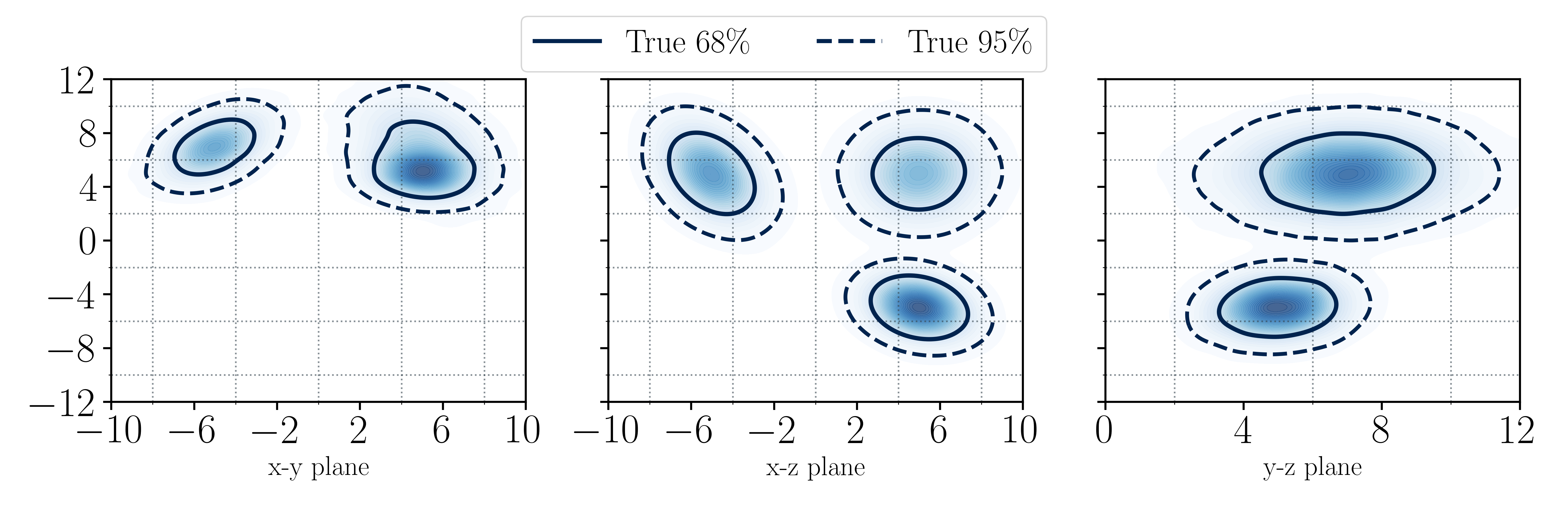}
\caption{\footnotesize Isosurface visualization of the ground truth 3-dimensional posterior distribution, displaying three distinct modes in the parameter space. The vertical axes share identical limits and tick spacing across all panels.}
\label{fig:3DTruePosterior}
\end{figure}

\begin{figure}[htbp]
\centering
\includegraphics[width=.99\textwidth, height = 5cm]{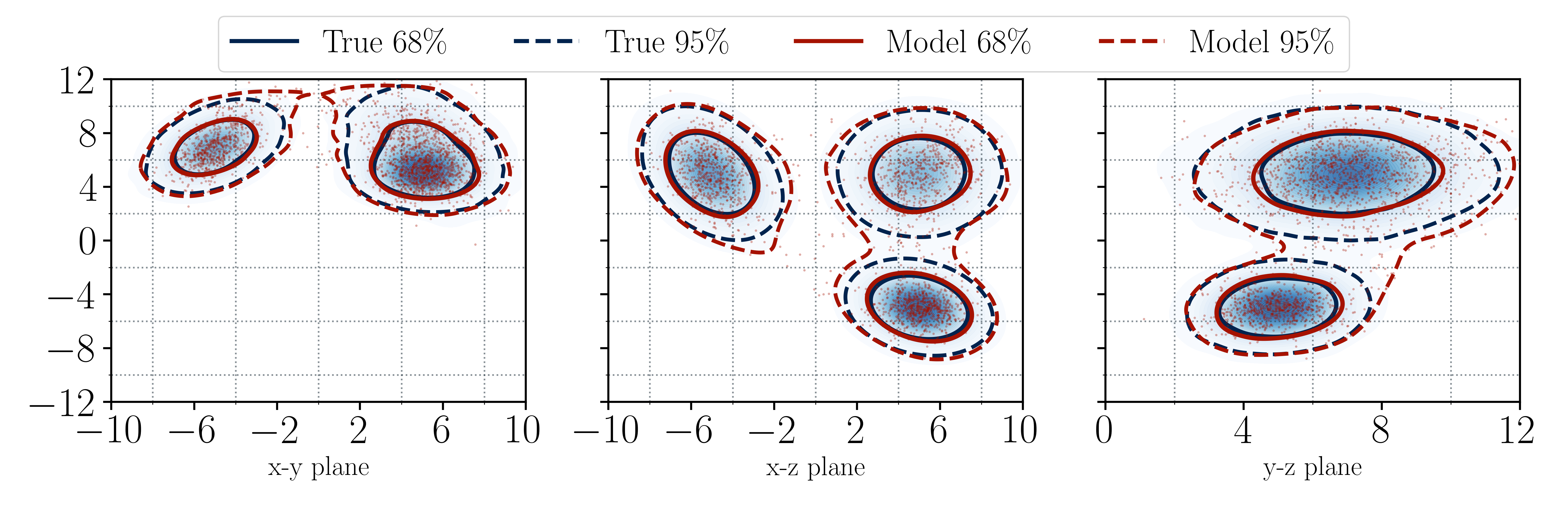}
\caption{\footnotesize 2D Marginal projections of the modelled 3D posterior. The flow captures the location of the modes, but leaves bridges between them with the unimodal base distribution. The vertical axes share identical limits and tick spacing across all panels}
\label{fig:3D_3ModePosteriorModelled}
\end{figure}

We train a RealNVP with a standard multivariate normal base distribution. The results, visualized via 2D marginal projections in \cref{fig:3D_3ModePosteriorModelled}, confirm the findings from the 2D case. Despite the higher dimensionality, the flow successfully locates all three modes. However, the same low-density bridges connect the distinct regions of high probability.

\section{Multi-modal Base Distributions}
\label{sec:Multimodality}

The fundamental objective of a Normalizing Flow is to construct a parameterized diffeomorphism between a tractable base distribution and a complex target density. In \cref{sec:implementation}, we demonstrated that a RealNVP network trained with likelihood weights successfully approximates the posterior. However, we also observed that a unimodal base distribution leaves bridges of probability between the modes. If the bridges come from the stretch that a flow needs to separate the modes, one remedy is to give the base separated regions of high density from the start. A Gaussian mixture with well-separated components provides them, so that the flow does not have to pull the modes apart itself. In this section, we systematically investigate the impact of the base distribution's structure. Specifically, we analyse how varying the number of modes in the latent space—while maintaining a fixed network architecture—affects the fidelity of the posterior reconstruction.

\subsection{Modality of the Base Distribution}
\label{sec:baseModality}
Building on the toy problems from \cref{sec:implementation}, we design an experiment to test the expressivity of multi-modal base distributions. We define a set of models, denoted as $\text{Model}-k\text{D}_i$, where $k$ represents the dimension of the parameter space and $i$ indicates the number of modes in the base distribution. Specifically, for both the 2-dimensional and 3-dimensional tasks, we evaluate flows initialized with base distributions consisting of 1, 2, and 3 equally weighted Gaussian components.

\begin{figure*}[t]
  \centering

\includegraphics[width=.99\textwidth, height = 5cm]{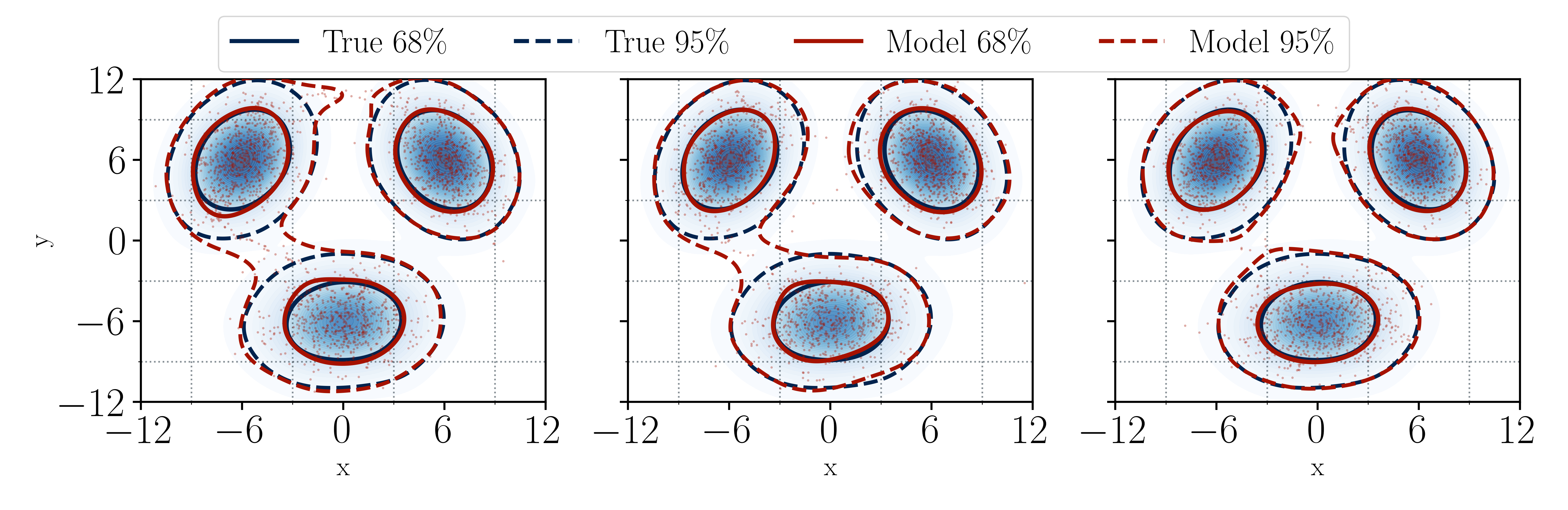}

  \caption{\footnotesize Comparison of learned posteriors from Normalizing Flows with different numbers of distinct modes in the base distribution.
  From Left to Right, the models have 1 base mode($Model-2D_1$), 2 base modes($Model-2D_2$) and 3 base modes($Model-2D_3$).}
  \label{fig:all_modes2D}
\end{figure*}

\begin{figure*}[t]
  \centering
\includegraphics[width=.99\textwidth, height = 10cm]{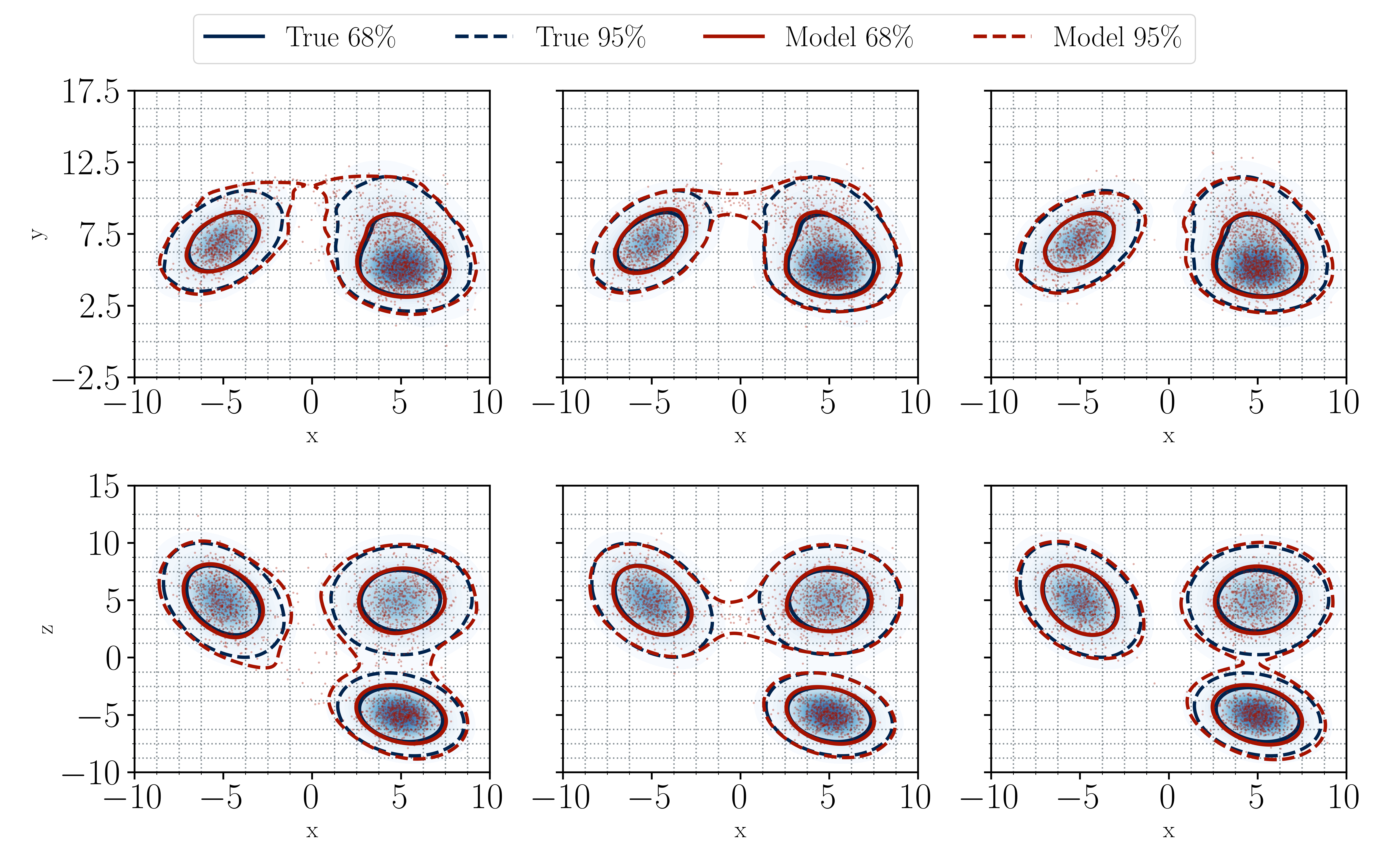}

  \caption{\footnotesize Comparison of learned posterior from Normalizing Flows with different no. of distinct modes in the base distribution. From Left to Right, the models have 1 base mode($Model-3D_1$), 2 base modes($Model-3D_2$) and 3 base modes($Model-3D_3$). The top and bottom rows are the projections on x-y plane and x-z plane, respectively.}\label{fig:all_modes}
\end{figure*}

The resulting modelled posteriors have been plotted in \cref{fig:all_modes2D,fig:all_modes} for the 2-D and 3-D toy problems, respectively. Comparing the plots, the modelled samples from $Model-2D_1$ 
 and $Model-3D_1$ \cref{fig:all_modes2D,fig:all_modes} clearly show the artifact of connected modes. In contrast, samples from $Model-2D_2$ 
 and $Model-3D_2$ \cref{fig:all_modes2D,fig:all_modes} exhibit partial separation, while $Model-2D_3$ 
 and $Model-3D_3$ \cref{fig:all_modes2D,fig:all_modes} achieve almost complete disconnection. In these tests, whenever the base has fewer regions of high density than the target has modes, the flow has to separate at least two modes by stretching, and it keeps the tails that connect them. 
One thing to keep in mind is that, although the models are different, they are fitted to the same weighted sample, and the final values of the loss are almost identical for all of them. The loss alone does not reveal the bridges, since the forward divergence penalises them only weakly; the distance metrics do.

When the number of components in the base matches the number of modes of the true posterior, all three metrics are the best. From \cref{tab:posterior_metrics_multiModal2D}, \(\text{Model}-2\text{D}_3\) reduces the KL divergence by a factor of almost three, the marginal Wasserstein distance by about $40\%$, and the sliced Wasserstein distance by about a quarter with respect to the Gaussian base\footnote{In practice, if the number of modes in the base distribution and the true posterior are high, then reaching an optimum loss while training is not very trivial in higher dimensions.}. The intermediate case is instructive. Two components improve the KL divergence but make both Wasserstein distances worse. A plausible explanation is that, with two components for three modes, one component has to be split between two modes, and the mass it leaves between them is further from where it belongs than the diffuse tails of a single Gaussian. So, adding components is not automatically an improvement. We note the limits of these tests. They cover at most three modes, and we did not test a base with more components than modes. They show that a base matched to the number of modes works best among the bases we tried, not that the match is necessary.

\begin{table}[h!]
\centering
\small
\begin{tabular}{lcccc}
\hline
\textbf{Model} & $\overline W_{1}$ & $\overline{\mathrm{SW}}_{1}$
 & $D_{\text{\small KL}}$ & $\overline W_{1}$ (floor) \\
\hline
\( \text{Model-2D}_1 \) & 0.1352 & 0.1291 & 0.0358 & 0.1070 \\
\( \text{Model-2D}_2 \) & 0.2372 & 0.2797 & 0.0216 & 0.1367 \\
\( \text{Model-2D}_3 \) & \textbf{0.0787} & \textbf{0.0948} & \textbf{0.0124} & 0.0914 \\
\hline
\end{tabular}
\caption{\small Performance comparison between different 2D models when the true posterior has three modes. Model-2D$_3$, whose number of base components matches the number of modes of the true posterior, is the best on all three metrics; its $\overline W_1$ is at the noise floor, i.e.\ its reconstruction is statistically indistinguishable from the truth at this sample size.}
\label{tab:posterior_metrics_multiModal2D}
\end{table}

\subsection{Non-Gaussian Distribution}
\label{sec:nonGauss}
\begin{figure*}[t]
  \centering

\includegraphics[width=.99\textwidth,                  height = 10cm]{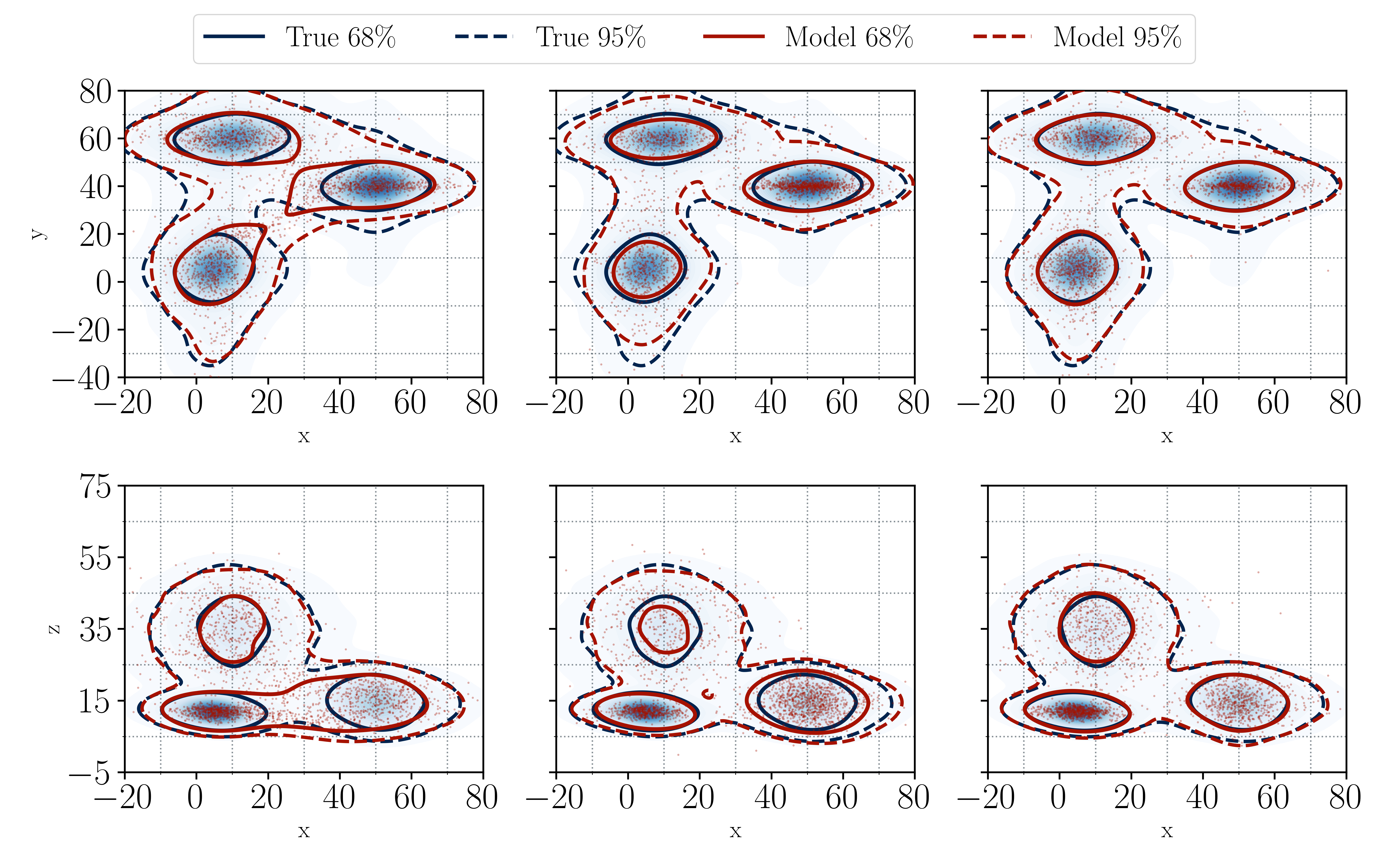}

  \caption{\footnotesize Comparison of learned posterior from Normalizing Flows with different no. of distinct modes in the base distribution, when the true posterior is a non-Gaussian distribution. From Left to Right, the models have 1 base mode(Model-nonGauss$_1$), 2 base modes(Model-nonGauss$_2$) and 3 base modes(Model-nonGauss$_3$). The top and bottom rows are the projections on x-y plane and x-z plane, respectively. }\label{fig:all_modesPath}
\end{figure*}

The posterior distributions considered so far are different Gaussians, either a single Gaussian or a mixture of Gaussians. In this subsection, we will look into a distribution in 3-dimensions which is a product of different non-Gaussian distributions. Given that the true posteriors are now non-Gaussian in nature and the base distributions are a mixture of Gaussians, this poses an interesting situation. We can see in \cref{fig:all_modesPath} how the modelled posterior compares with the true posterior for three different models, where, Model-nonGauss$_i$ corresponds to the base distribution having \(i\) distinct modes. The distance and divergence metrics for the three models are shown in  \cref{tab:posterior_metrics_multiModal3D_Path}  and again the base whose number of components matches the number of modes gives the best metrics. However, the agreement is worse than in the Gaussian case. The Wasserstein distances depend on the scale of the target and cannot be compared directly between the two problems, but the KL divergence, which does not, is also larger, by a factor of about $2.6$ for the best base. We conclude that the method estimates Gaussian posteriors more accurately than non-Gaussian ones. A base with the right number of modes improves the result, but it does not remove the difficulty of a non-Gaussian shape.

\begin{table}[ht!]
\centering
\small
\begin{tabular}{lcccc}
\hline
\textbf{Model} & $\overline W_{1}$ & $\overline{\mathrm{SW}}_{1}$
 & $D_{\text{\small KL}}$ & $\overline W_{1}$ (floor) \\
\hline
Model-nonGauss$_1$ & 1.6173 & 1.5800 & 0.1699 & 0.3340 \\
Model-nonGauss$_2$ & 4.4237 & 4.4310 & 0.0722 & 0.3517 \\
Model-nonGauss$_3$ & \textbf{0.3732} & \textbf{0.3788} & \textbf{0.0322} & 0.2291 \\
\hline
\end{tabular}
\caption{\small Performance comparison between the three 3D models when the true posterior is non-Gaussian with three modes. Model-nonGauss$_3$ is again the best on every metric, matching the trend of \cref{tab:posterior_metrics_multiModal2D}; the overall performance is, however, poorer than in the Gaussian case, and all three sit above the noise floor. The reference sample for this target contains a substantial fraction of repeated rows (it was generated by resampling with replacement), which inflates the effective precision of density-based estimators; the $\overline W_1$ and $\overline{\mathrm{SW}}_1$ columns are unaffected.}
\label{tab:posterior_metrics_multiModal3D_Path}
\end{table}

All entries in \cref{tab:metrics_2D_singlebase,tab:posterior_metrics_multiModal2D,tab:posterior_metrics_multiModal3D_Path} are computed between the modelled samples and the reference posterior with the script released alongside this paper. $\overline W_1$ is the average overdimensions of the one-dimensional Wasserstein-1 distance (\cref{app:w1}), $\overline{\mathrm{SW}}_1$ is the sliced Wasserstein distance of \cref{app:sw} evaluated over $10^3$ random projections, and $D_{\text{\small KL}}$ is a Monte-Carlo estimate of $\mathbb{E}_p[\log p - \log q]$ using Gaussian kernel density estimates, with the reference sample split in half so that the density is never evaluated on the points used to fit it. Because a sample-based KL is estimator-dependent, the released script also reports the $k$-nearest-neighbour estimator of Perez-Cruz as a cross-check; the two agree on the ordering of the models in every case. $\overline{\mathrm{SW}}_1$ carries a Monte-Carlo uncertainty of order $10^{-3}$ from the random projections.

\section{B Meson CP Violation: A Multimodal, Non-Gaussian Benchmark}
\label{sec:bmeson}

We use the time-dependent CP asymmetry in $B^0 \to J/\psi\, K^0$ as a physically motivated case for posterior inference and compare the results acquired using this method with a standard MCMC. The problem is chosen deliberately for two reasons. First, it is \emph{genuinely multimodal}. The symmetry $\sin(2\beta)=\sin(\pi-2\beta)$ produces two well-separated solutions in the Wolfenstein plane. Any method that reports a single mode, or that misweights the two, has mischaracterised the posterior. Second, it is strongly \emph{non-Gaussian}. The CP asymmetry constraint confines $(\bar\rho,\bar\eta)$ to two thin wedges that open up from the vertex $(1,0)$, the amplitude enters through a product degeneracy $|V_{cb}|g$, and the prior is uniform in parameters that are nonlinearly related to the physical angle $\beta$. A correct posterior is therefore neither unimodal nor elliptical; it is not even symmetric in weight between its two modes. This makes it an ideal benchmark for likelihood-weighted normalising flows (LW-NF), whose value proposition is precisely the ability to represent multimodal, elongated densities. A third property makes it a good benchmark: its posterior factorises and can be computed exactly by quadrature (\cref{sec:bmeson:exact}), so the answer is known to high precision before any method is run.

We study the four-dimensional parametrisation, which is the richest of the family and the one that combines all of the above difficulties.

\subsection{Physical Background}
\label{sec:bmeson:physics}

The time-dependent CP asymmetry in $B^0 \to J/\psi\, K^0$ is one of the cleanest probes of the Cabibbo--Kobayashi--Maskawa (CKM) matrix~\cite{Cabibbo:1963yz,Kobayashi:1973fv}. In the Wolfenstein
parametrisation~\cite{Wolfenstein:1983yz} the matrix is expressed through $(\lambda,\,A,\,\bar\rho,\,\bar\eta)$, with $(\bar\rho,\,\bar\eta)$ being the apex of the Unitarity Triangle. The CP-violating angle $\beta$ is
\begin{equation}
    \beta = \arctan\left(\frac{\bar\eta}{1-\bar\rho}\right),
    \label{eq:beta_def}
\end{equation}
and the measured asymmetry is
\begin{equation}
    \mathcal{A}_{CP}(t) = -\sin(2\beta)\,\sin(\Delta m_d\,t),
    \qquad
    \sin(2\beta) = \frac{2\bar\eta\,(1-\bar\rho)}{(1-\bar\rho)^2+\bar\eta^2}.
    \label{eq:sin2b}
\end{equation}
Since $\mathcal{A}_{CP}$ is a ratio of rates, the hadronic amplitude $|A|=|V_{cb}|\cdot g$ cancels and the asymmetry constrains $\sin(2\beta)$ independently of $|A|$. The amplitude is fixed separately by the branching fraction $\mathcal{B}(B^0\to J/\psi\,K^0)=(8.73\pm 0.32)\times10^{-4}$~\cite{PDG:2024}, which is proportional to $|A|^2$; normalising
$|A|=1$ at the central value gives $|A|^2 = 1.000\pm 0.037$ with
$\sigma_{\mathcal{B}}=0.32/8.73=0.037$. The 4D problem additionally uses the direct exclusive determination from $B\to D^{\ast}\ell\nu$,
\begin{equation}
    |V_{cb}|^{\rm obs} = (39.8\pm 0.6)\times 10^{-3}
    \quad\text{\cite{PDG:2024}},
    \label{eq:vcb_obs}
\end{equation}
together with the CP-asymmetry measurement
\begin{equation}
    \mathcal{A}_{CP}^{\rm obs} = 0.700 \pm 0.017
    \quad\text{\cite{HFLAV:2023}}.
    \label{eq:acp_obs}
\end{equation}

\begin{figure*}[htb]
    \centering

    \begin{subfigure}[t]{0.48\textwidth}
        \centering
        \includegraphics[width=\linewidth]{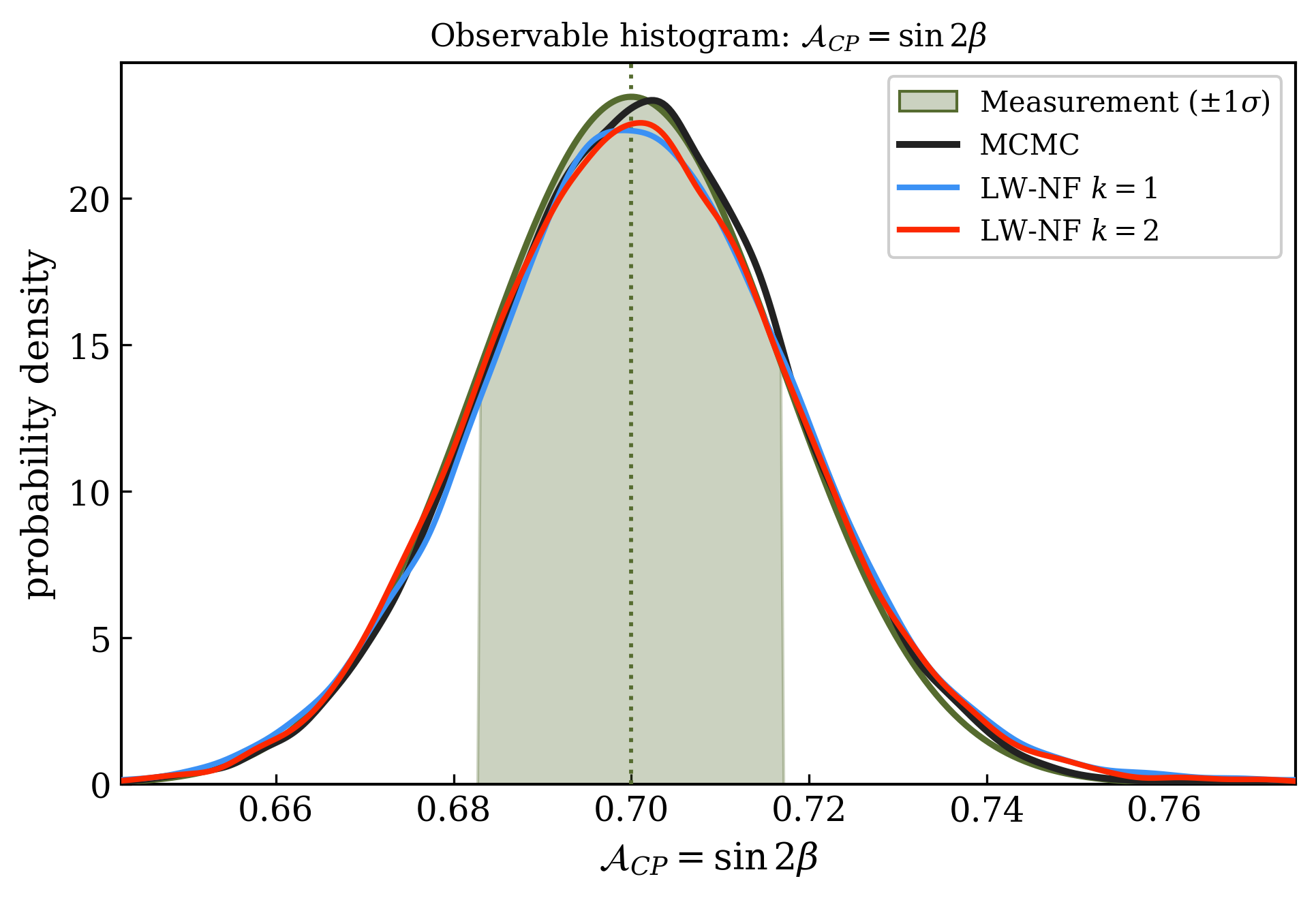}
        \caption{\footnotesize
    Marginal posterior in the derived observable
    $\mathcal{A}_{CP}=\sin 2\beta$. The shaded green band denotes the experimental $1\sigma$ interval centred on the measured value (vertical dotted line). The MCMC reference posterior (black) and the likelihood-weighted normalizing flows with $k{=}1$ (blue) and $k{=}2$ (red) yield nearly identical predictive distributions, consistent with the measurement. Since both modes predict the same asymmetry, this does not test the weights of the modes (see text).}
\label{fig:obs_acp_hist}
    \end{subfigure}
    \hfill
    \begin{subfigure}[t]{0.48\textwidth}
        \centering
        \includegraphics[width=\linewidth]{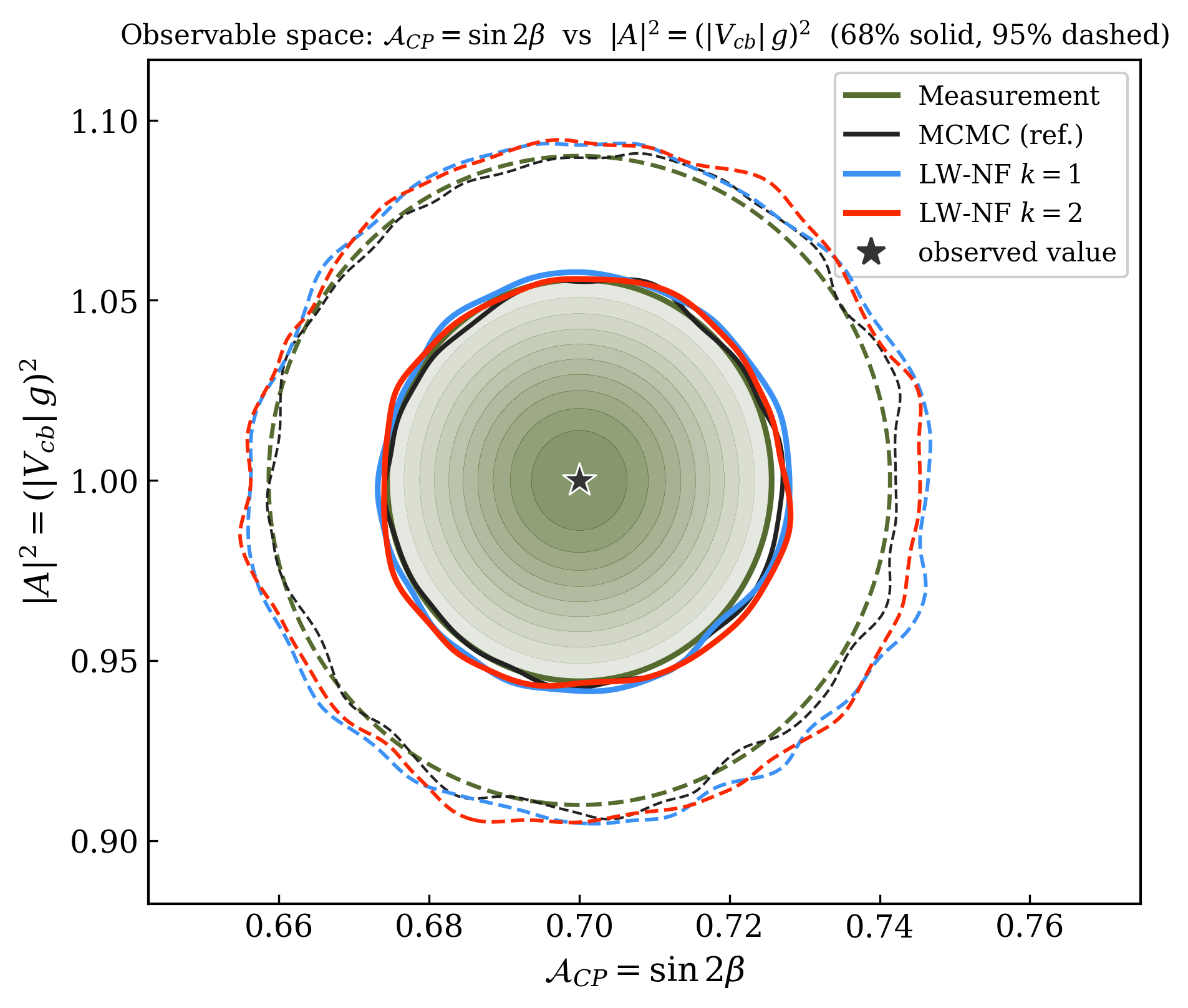}
        \caption{\footnotesize
    Posterior credible regions in the observable plane
    $(\mathcal{A}_{CP},|A|^2)$, where
    $|A|^2=(|V_{cb}|\,g)^2$. Solid and dashed contours denote the
    $68\%$ and $95\%$ credible regions, respectively. The shaded green
    contours represent the measurement likelihood, while the star marks
    the observed central value. The MCMC reference (black) and the
    likelihood-weighted normalizing flows with $k{=}1$ (blue) and
    $k{=}2$ (red) show excellent agreement in observable space.}
\label{fig:obs_acp_a2}
    \end{subfigure}

    \caption{\footnotesize The probability density plot of the observables in 1-D($\mathcal{A}_{CP}=\sin 2\beta$) and 2-D($(\mathcal{A}_{CP},|A|^2)$)}
    \label{fig:obs}
\end{figure*}

The figures \cref{fig:obs_acp_hist,fig:obs_acp_a2} show the observable distributions; both the actual experimental distributions and the calculated distributions from the posteriors measured using different methods. In this space the two modes fall on the same region, since both reproduce the measured asymmetry, and the distributions predicted by the flows are nearly identical to those of the reference. This shows that the flows have learnt where the likelihood is large, but it is a weak test of the posterior itself: a posterior with any split between the modes, including $50/50$, would give the same distribution of $\mathcal{A}_{CP}$. We can check the weights of the modes only in parameter space (\cref{sec:results}).

\subsection{Parametrisation and Degeneracies}
\label{sec:bmeson:param}

The 4D model treats $\theta=(\bar\rho,\,\bar\eta,\,g)$ as free parameters and $|V_{cb}|$ as a nuisance parameter. The amplitude enters as $|A|=|V_{cb}|\,g$.

Since $\sin(2\beta)=\sin(\pi-2\beta)$, both $\beta_1$ and $\beta_2=\pi/2-\beta_1$ reproduce the same $\mathcal{A}_{CP}$. Numerically $\beta_1\approx 0.388$~rad and $\beta_2\approx 1.183$~rad, which at the true $\bar\eta=0.350$ map to two values of $\bar\rho$,
\begin{equation}
    \bar\rho_1 \approx 0.143
    \qquad\text{and}\qquad
    \bar\rho_2 \approx 0.857,
    \qquad
    \Delta\bar\rho \approx 0.714 
    \label{eq:two_rho}
\end{equation}
The two solutions are separated by a wide region in which $\sin(2\beta)$
departs strongly from the measured value, so the likelihood between them is
vanishingly small: the modes are isolated, not connected by any ridge of
support.

The amplitude constraint $|A|^2=1$ fixes only the product $|V_{cb}|\,g$,
defining a hyperbola $g=1/|V_{cb}|$ in the $(|V_{cb}|,g)$ subspace. The
direct measurement of \cref{eq:vcb_obs} breaks this degeneracy, pinning
$|V_{cb}|\approx 0.0398$ and hence $g\approx 1/0.0398\approx 25.13$. The
$(|V_{cb}|,g)$ marginal is therefore unimodal; the two physical modes remain distinguished \emph{only} by $\bar\rho$.

For fixed $\mathcal{A}_{CP}$, \cref{eq:sin2b} is one equation in two unknowns. Each solution for $\beta$ is a straight line through the vertex $(1,0)$, $1-\bar\rho = \bar\eta/\tan\beta_m$, and with the experimental uncertainty each line becomes a narrow wedge that opens up away from the vertex. Along each wedge, $\bar\eta$ is constrained only by the prior. The posterior is thus a pair of thin, elongated structures rather than a pair of isolated peaks, and $\bar\eta$ is the soft direction along them. This is the geometric source of the non-Gaussianity.

\subsection{Likelihood and Priors}
\label{sec:bmeson:like}

We use a Gaussian likelihood $\ell(\theta)=-\tfrac12\chi^2(\theta)$ with
\begin{equation}
    \chi^2(\theta)
    =
    \left(\frac{\sin(2\beta)-\mathcal{A}_{CP}^{\rm obs}}{\sigma_{\mathcal{A}}}\right)^{\!2}
    +
    \left(\frac{(|V_{cb}|\,g)^2-1}{\sigma_{\mathcal{B}}}\right)^{\!2}
    +
    \left(\frac{|V_{cb}|-|V_{cb}|^{\rm obs}}{\sigma_V}\right)^{\!2},
    \label{eq:like4d}
\end{equation}
where $\sin(2\beta)$ is evaluated through \cref{eq:sin2b}, and $\sigma_{\mathcal{A}}=0.017$, $\sigma_{\mathcal{B}}=0.037$, $\sigma_V=6\times 10^{-4}$. The first term constrains the angle (and is what generates the two modes), the second enforces the unit amplitude, and the third resolves the product degeneracy. The prior is a uniform hyper-cuboid (\cref{tab:prior_bounds}) and also it is uniform in $(\bar\rho,\bar\eta)$, \emph{not} in $\beta$, a choice with direct consequences for the mode weights (\cref{sec:bmeson:expect}).

The structure of \cref{eq:like4d} has a useful consequence. The first term depends only on $(\bar\rho,\bar\eta)$, and the other two only on $(|V_{cb}|,g)$. Since the prior is a product of uniform distributions, the posterior factorises,
\begin{equation}
    p(\theta\,|\,D) = p(\bar\rho,\bar\eta\,|\,\mathcal{A}_{CP}^{\rm obs})\;
                      p(|V_{cb}|,g\,|\,\mathcal{B},|V_{cb}|^{\rm obs}),
    \label{eq:factorise}
\end{equation}
and the evidence is the product of the evidences of the two blocks. The exact posterior therefore follows from two two-dimensional integrals (\cref{sec:bmeson:exact}). We do not tell the flows about this factorisation; we train them in four dimensions, as we would for a problem without this structure.

\begin{table}[htb]
\centering
\begin{tabular}{@{}lcccc@{}}
\toprule
Parameter & $\bar\rho$ & $\bar\eta$ & $|V_{cb}|$ & $g$ \\
\midrule
Prior & $[-0.10,\,0.97]$ & $[0.20,\,0.50]$ & $[0.020,\,0.070]$ & $[10,\,40]$ \\
\bottomrule
\end{tabular}
\caption{\small Uniform prior bounds for the 4D problem. The $\bar\rho$ range
    contains both solutions, $\bar\rho_1\approx0.143$ and
    $\bar\rho_2\approx0.857$.}
\label{tab:prior_bounds}
\end{table}

\subsection{Theoretical Expectations}
\label{sec:bmeson:expect}

Before turning to inference, we state what the true posterior should look like, since the benchmark value of the exercise is that this is known.

Both solutions satisfy $\sin(2\beta)=0.700$ exactly, so they share the same peak likelihood. It may seem obvious to conclude that each mode carries $50\%$ of the posterior \textit{mass}. This is \emph{not} the case here, because the prior is flat in $\bar\rho$, not in
$\beta$. We can derive the weights of the modes with a short argument. Consider a slice of fixed $\bar\eta$. The asymmetry constrains $\beta$ to a narrow interval $\delta\beta = \sigma_{\mathcal{A}}/(2|\cos 2\beta_m|)$ around each solution, and this interval is the same for both, since $|\cos 2\beta_2| = |\cos(\pi-2\beta_1)| = |\cos 2\beta_1|$. In $\bar\rho$, however, the two intervals differ. From \cref{eq:beta_def}, $\partial\beta/\partial\bar\rho = \bar\eta/[(1-\bar\rho)^2+\bar\eta^2]$, and on the line $1-\bar\rho_m = \bar\eta/\tan\beta_m$ this equals $\sin^2\beta_m/\bar\eta$. The width in $\bar\rho$ of solution $m$ is therefore
\begin{equation}
    \delta\bar\rho_m = \frac{\bar\eta}{\sin^2\beta_m}\,\delta\beta ,
    \qquad
    \frac{\delta\bar\rho_1}{\delta\bar\rho_2} = \frac{\sin^2\beta_2}{\sin^2\beta_1} = \frac{1}{\tan^2\beta_1} \approx 6.0 ,
    \label{eq:slopes}
\end{equation}
with $\tan\beta_1 = 0.408$. Since the likelihood has the same height on both solutions and the prior is flat in $\bar\rho$, the mass of each solution in the slice is proportional to its width. Mode~1 is flatter in $\bar\rho$, hence wider, hence heavier, by a factor of six at \emph{every} value of $\bar\eta$. This is the ratio of the slopes $|\partial\sin(2\beta)/\partial\bar\rho|$ of the two solutions, $3.50/0.583$ at $\bar\eta=0.350$; \cref{eq:slopes} shows that it does not depend on $\bar\eta$.

A ratio of six in every slice would give $f_1 = 6/7 \approx 0.86$. The true fraction is smaller for a different reason, which is the boundary of the prior. \Cref{eq:slopes} also shows that both widths grow linearly with $\bar\eta$, since the wedges open up away from $(1,0)$, so in the interior of the prior the mass of each mode per unit $\bar\eta$ is proportional to $\bar\eta$. The second solution, $\bar\rho_2 = 1-\bar\eta\tan\beta_1$, lies between $0.80$ and $0.92$ for every $\bar\eta$ in the prior and is never cut. The first, $\bar\rho_1 = 1-\bar\eta/\tan\beta_1$, reaches the lower edge $\bar\rho=-0.10$ at $\bar\eta_{\rm w} = 1.10\tan\beta_1 \approx 0.449$, beyond which it leaves the prior. In the limit of a narrow constraint, the ratio of the masses is therefore
\begin{equation}
    \frac{f_1}{f_2} \;\approx\; \frac{1}{\tan^2\beta_1}\,
    \frac{\int_{0.20}^{\bar\eta_{\rm w}}\bar\eta\,{\rm d}\bar\eta}{\int_{0.20}^{0.50}\bar\eta\,{\rm d}\bar\eta}
    \;\approx\; 6.0\times\frac{0.081}{0.105} \;\approx\; 4.6
    \quad\Longrightarrow\quad
    f_1 \approx 0.822 .
    \label{eq:weight_ratio}
\end{equation}
The exact posterior (\cref{sec:bmeson:exact}) gives $f_1 = 0.8220$ when each point is assigned to the solution on which it lies, that is, by $\beta < \pi/4$. In the results below, we split the samples instead at the midpoint $\bar\rho = 0.5$ between the modes (\cref{app:mode}). This assigns to mode~2 the part of the first solution with $\bar\eta < 0.5\tan\beta_1 \approx 0.204$, where $\bar\rho_1 > 0.5$, together with a little of the width of the constraint, and the exact value becomes $f_1 = 0.8111$ ($f_1/f_2 \approx 4.3$). Both definitions are legitimate, but they must not be mixed. The essential point is that the true 4D posterior is strongly asymmetric, $\approx 81\%/19\%$ by the midpoint split, and a faithful method must reproduce this imbalance, not a $50/50$ split.

We therefore expect (a) a bimodal $\bar\rho$ marginal, with solutions at $\bar\rho\approx0.143$ and $0.857$ for the injected $\bar\eta=0.350$, carrying $\approx81\%$ and $19\%$ of the mass; (b) a broad, non-Gaussian $\bar\eta$ marginal, which rises linearly, peaks at $\bar\eta\approx0.42$ just before the first solution reaches the boundary of the prior, and then falls by a factor of about five by $\bar\eta=0.48$, so that its peak has nothing to do with the injected $\bar\eta=0.350$; (c) a narrow, near-Gaussian $|V_{cb}|$ marginal at $0.0398\pm0.0006$ fixed by the direct measurement; and (d) a unimodal $g$ marginal centred near $25.13$, the image of the broken product degeneracy. These four expectations are the yardstick against which the inference methods are judged.

\subsection{Inference Methods}
\label{sec:methods}

\subsubsection{Reference Posterior: Markov Chain Monte Carlo}
\label{sec:methods:mcmc}

We take a long, well-converged Markov Chain Monte Carlo run as the reference posterior. We use the \texttt{OptEx} package in Mathematica~\cite{sunando_patra_2019_3404311} with a single walker and a large spread(which was decided upon by a preliminary nested sampling run), so that both $\bar\rho$ lobes are populated, and run far beyond autocorrelation convergence; the flat chain after burn-in is retained as the reference. The resulting reference mode fraction, $f_1=0.809$, agrees with the exact value $f_1=0.8111$ (\cref{sec:bmeson:exact}), which validates the chain as a trustworthy reference for the asymmetric posterior. The distance metrics below are quoted relative to this MCMC reference, except the truth-based Mahalanobis distance (\cref{app:mahal_truth}), which is measured against the \emph{injected}
parameter values.

\subsubsection{Exact Reference: Quadrature}
\label{sec:bmeson:exact}

Because of the factorisation of \cref{eq:factorise}, the evidence, the mode fractions, and the marginals of the posterior follow from two two-dimensional integrals over the prior box. We compute them with composite Gauss--Legendre quadrature, with 40 nodes in each panel and 400 and 60 panels in $\bar\rho$ and $\bar\eta$. Changing the number of panels and nodes changes $\ln Z$ by less than $10^{-6}$ and the mode fractions by less than $10^{-4}$. The exact posterior gives $f_1 = 0.8111$ with the midpoint split and $f_1 = 0.8220$ with the split at $\beta=\pi/4$, and $\ln Z = -9.4989$, up to the normalisation constant of the Gaussian likelihood. The exact means of the two modes are $(\bar\rho,\bar\eta) = (0.162, 0.343)$ and $(0.829, 0.362)$. They are displaced from the injected solutions, $(0.143, 0.350)$ and $(0.857, 0.350)$, by the prior, so the truth-based Mahalanobis distance of the exact posterior is not zero, but $d^{(1)}_{\rm TrMah} = 0.124$ and $d^{(2)}_{\rm TrMah} = 0.364$. A faithful method has to reproduce these values, not zero. We keep the MCMC chain as the reference for the distance metrics, which need samples.

\subsubsection{Likelihood-Weighted Normalising Flows}
\label{sec:methods:nf}

A normalising flow $f_\theta\colon\mathbb{R}^d\to\mathbb{R}^d$ maps a tractable
base $p_0(z)$ to the target. We minimise the importance-weighted negative
log-likelihood
\begin{equation}
    \mathcal{L}(\theta)
    = -\,\mathbb{E}_{x\sim q}\!\left[w(x)\,\log p_\theta(x)\right],
    \qquad
    w(x)\propto e^{\ell(x)},
    \label{eq:wnll}
\end{equation}
where $q$ is the uniform prior. Training points are drawn uniformly over the prior box; the $\sim\!2\times10^5$ points with non-negligible weight
($w>w_{\max}\times10^{-8}$) are retained, parameters are standardised by the prior box, and weights are clipped at the $99.9$th percentile to stop a handful of high-likelihood points from dominating the gradient. We need about $4.5\times10^7$ prior draws to collect these points, and the effective sample size of the training set is $N_{\rm eff}\approx 9.5\times10^3$ before clipping and $9.6\times10^3$ after. By \cref{eq:ess_chi2}, only about two in $10^4$ prior draws are effectively used.

Each coupling block applies, in sequence,(i)~\textbf{ActNorm}~\cite{Kingma:2018}, a per-channel affine normalisation, (ii)~a \textbf{LU-decomposed linear permutation}~\cite{Hoogeboom:2019}, a learned full-rank mixing layer, and (iii)~an \textbf{affine coupling} whose log-scale is clamped as $s=\tanh(s_{\rm raw})\times c$ to prevent log-determinant divergence~\cite{Stimper:2023}. This bound also limits how strongly each block can stretch the space, which is the limitation discussed in \cref{sec:implementationandresults}. Training uses Adam~\cite{Kingma:2015} with cosine learning-rate annealing and early stopping on a held-out validation loss.

We compare two bases. The \emph{Gaussian base} ($k{=}1$) is a single diagonal $\mathcal{N}(0,I_d)$: the flow must deform one connected blob into a disconnected, asymmetric bimodal target. The \emph{GMM-2 base} ($k{=}2$) is a two-component Gaussian mixture with trainable means, scales and weights, preceded by a short warm-up that trains the base alone before the coupling layers are activated. Importantly, the two components are initialised symmetrically along the diagonal of standardised space, \emph{not} at the true mode locations: the GMM-2 base supplies two regions of high density without injecting any prior knowledge of where the modes sit. Their separation is learned from the weighted data like everything else.

\paragraph{Ensemble protocol and uncertainty}
A single flow's quality depends on its random initialisation and on the
chosen architecture. To report a robust posterior with a genuine uncertainty, we run, for each base, a sweep of ten architecture configurations (varying depth, hidden width and the scale clamp $c$), train each once, and retain the \textbf{five with the lowest validation loss}. All five cleared the training target (best validation losses in the range $-3.08$ to $-2.96$ for $k{=}1$ and $-3.00$ to $-2.91$ for $k{=}2$). The reported flow posterior is the average over these five runs, and the shaded bands in all figures are the run-to-run $\pm1\sigma$ spread, which quantifies the residual sensitivity to
initialisation and architecture. This spread measures how much the flows
differ from each other, which is not the same as how far they are from the
truth (\cref{sec:results}). Posterior samples are drawn by a single
forward pass through each trained flow.

\subsection{Results}
\label{sec:results}

\begin{figure*}[htb]
    \centering

    \begin{subfigure}[t]{0.48\textwidth}
        \centering
        \includegraphics[width=\linewidth]{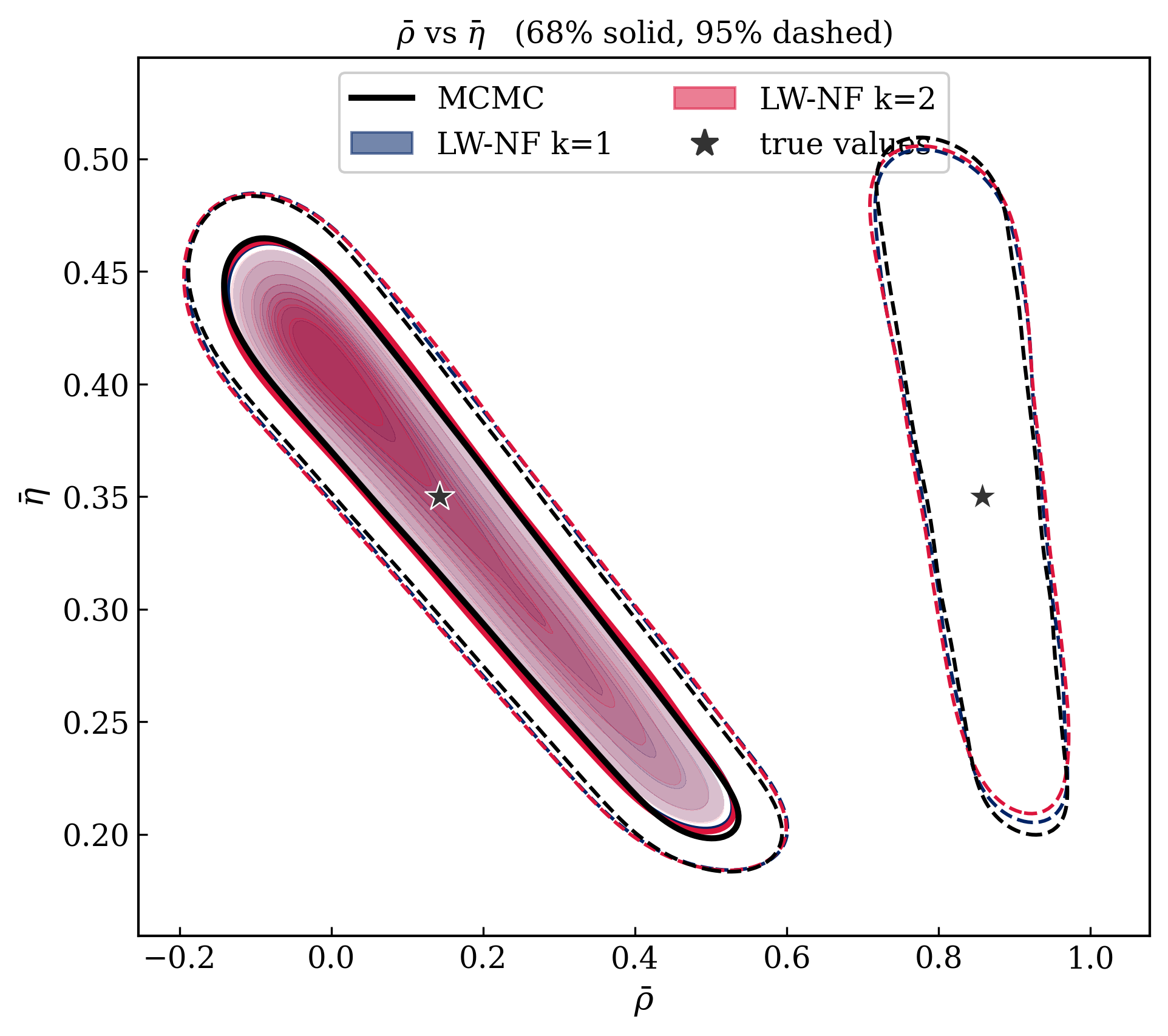}
        \caption{\footnotesize $(\bar\rho,\bar\eta)$ contours: MCMC $68/95\%$ levels (black) with the graded $k{=}1$ (navy) and $k{=}2$ (crimson) credible regions, pooled over the best five runs. The two isolated lobes, each elongated in $\bar\eta$, are recovered; stars mark the injected truth.}
        \label{fig:cont_rho_eta}
    \end{subfigure}
    \hfill
    \begin{subfigure}[t]{0.48\textwidth}
        \centering
        \includegraphics[width=\linewidth]{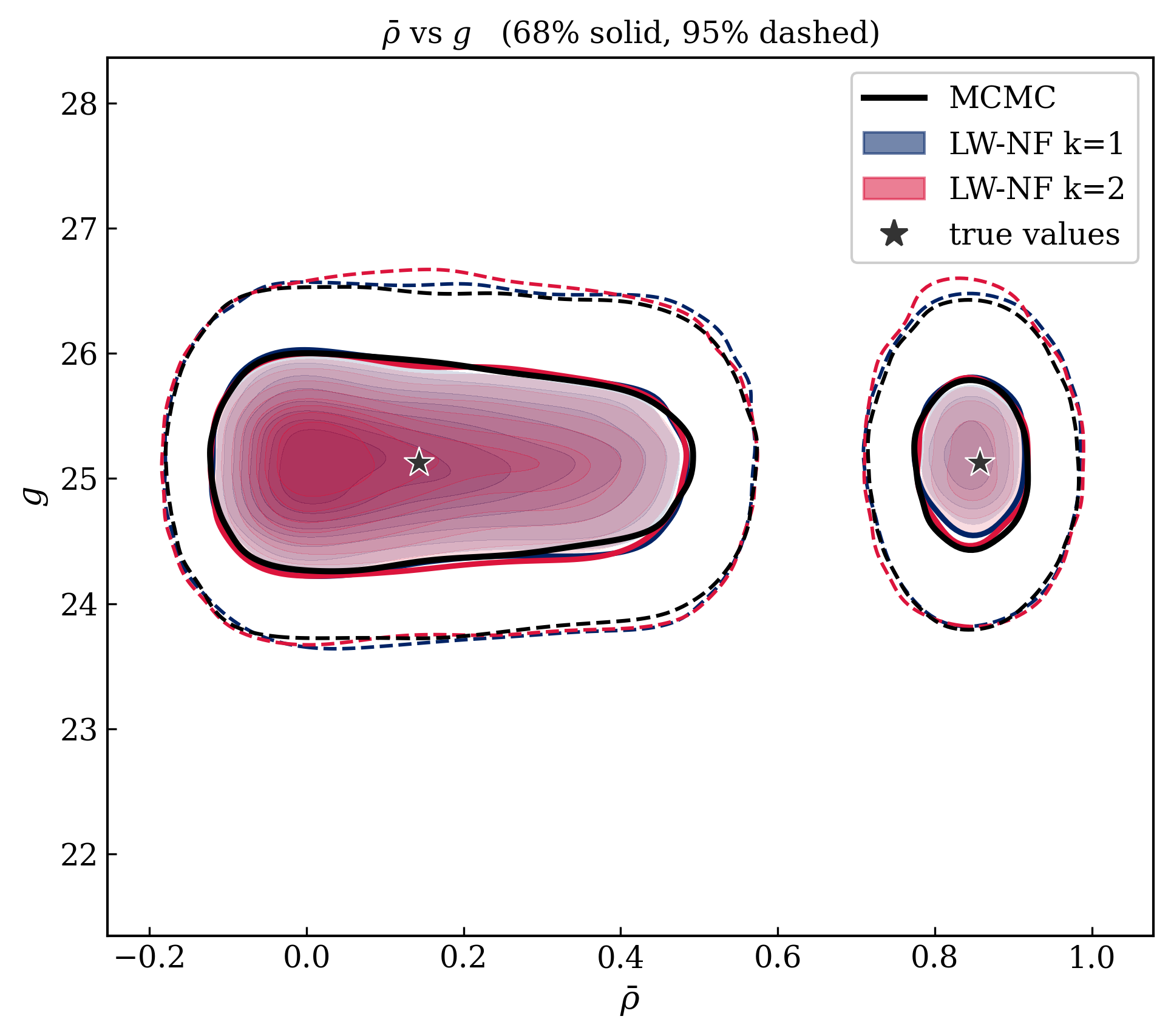}
        \caption{\footnotesize $(\bar\rho,g)$ contours. The bimodality in $\bar\rho$ persists while $g$ is unimodal about $g\approx25.13$, the image of the product degeneracy $|V_{cb}|\,g=1$ broken by the direct $|V_{cb}|$
        measurement. Both flow bases track the MCMC reference.}
        \label{fig:cont_rho_g}
    \end{subfigure}

    \caption{\footnotesize Two-dimensional credible regions from MCMC and likelihood-weighted normalizing flows in the $(\bar\rho,\bar\eta)$ and $(\bar\rho,g)$ planes.}
    \label{fig:contours}
\end{figure*}

\Cref{tab:mode_metrics,tab:posterior_metrics} collects the metrics, averaged over the best five runs for each base configuration. \Cref{fig:marg_rho,fig:marg_eta,fig:cont_rho_eta,fig:cont_rho_g} show the marginals and pairwise contours most relevant to the multimodal, non-Gaussian structure. 

{
\captionsetup[subfigure]{font=small}
\begin{figure*}[htb]
    \centering

    \begin{subfigure}[t]{0.48\textwidth}
        \centering
        \includegraphics[width=\linewidth]{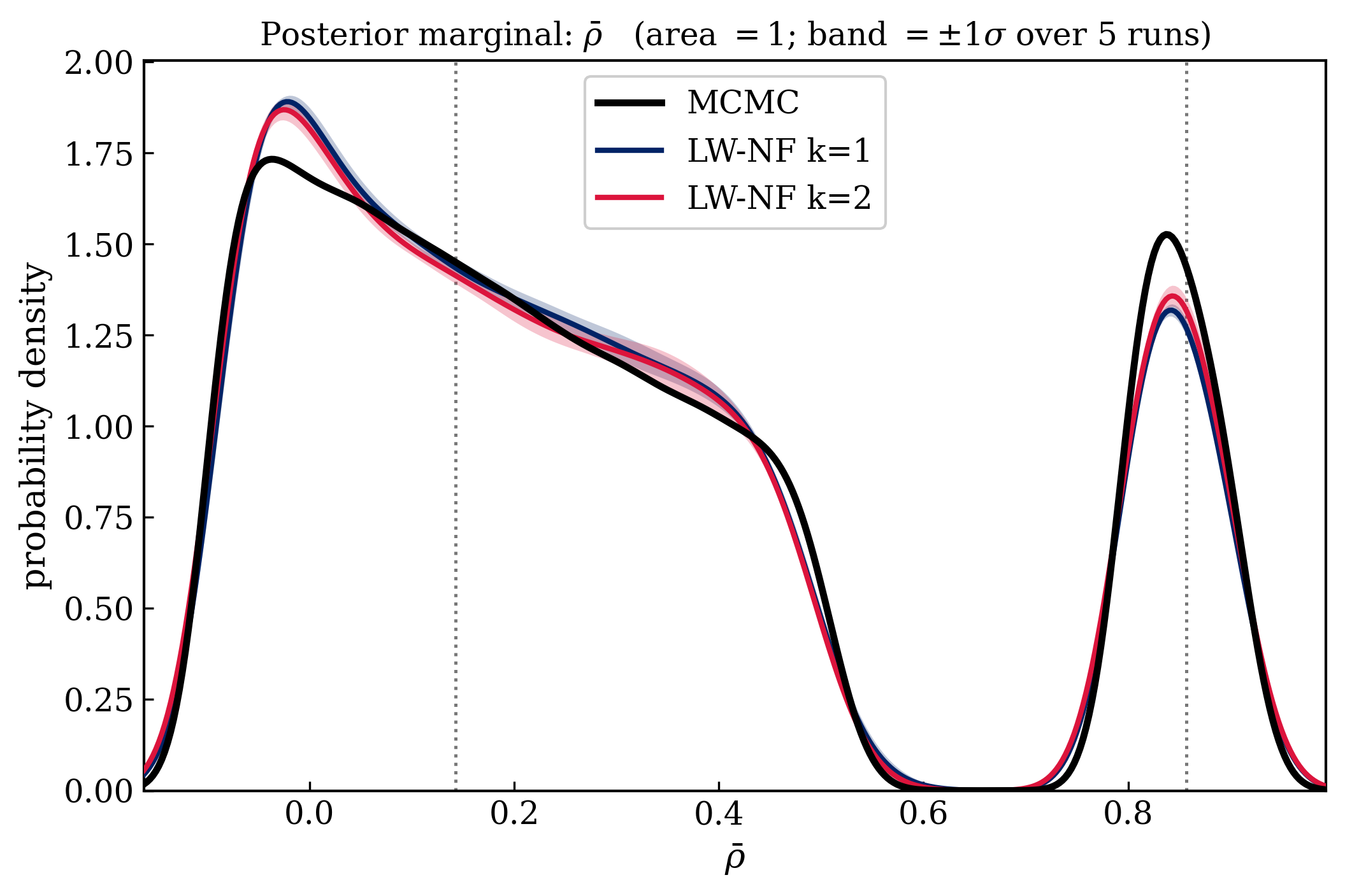}
        \caption{\footnotesize
        Marginal posterior in $\bar\rho$: MCMC reference (black) versus the
        LW-NF averages for $k{=}1$ (navy) and $k{=}2$ (crimson), with
        run-to-run $\pm1\sigma$ bands. Both bases recover the two modes at
        $\bar\rho\approx0.143$ and $0.857$ with very nearly the correct asymmetric weight
        ($\approx81/19$); the dotted lines mark the injected mode locations.}
        \label{fig:marg_rho}
    \end{subfigure}
    \hfill
    \begin{subfigure}[t]{0.48\textwidth}
        \centering
        \includegraphics[width=\linewidth]{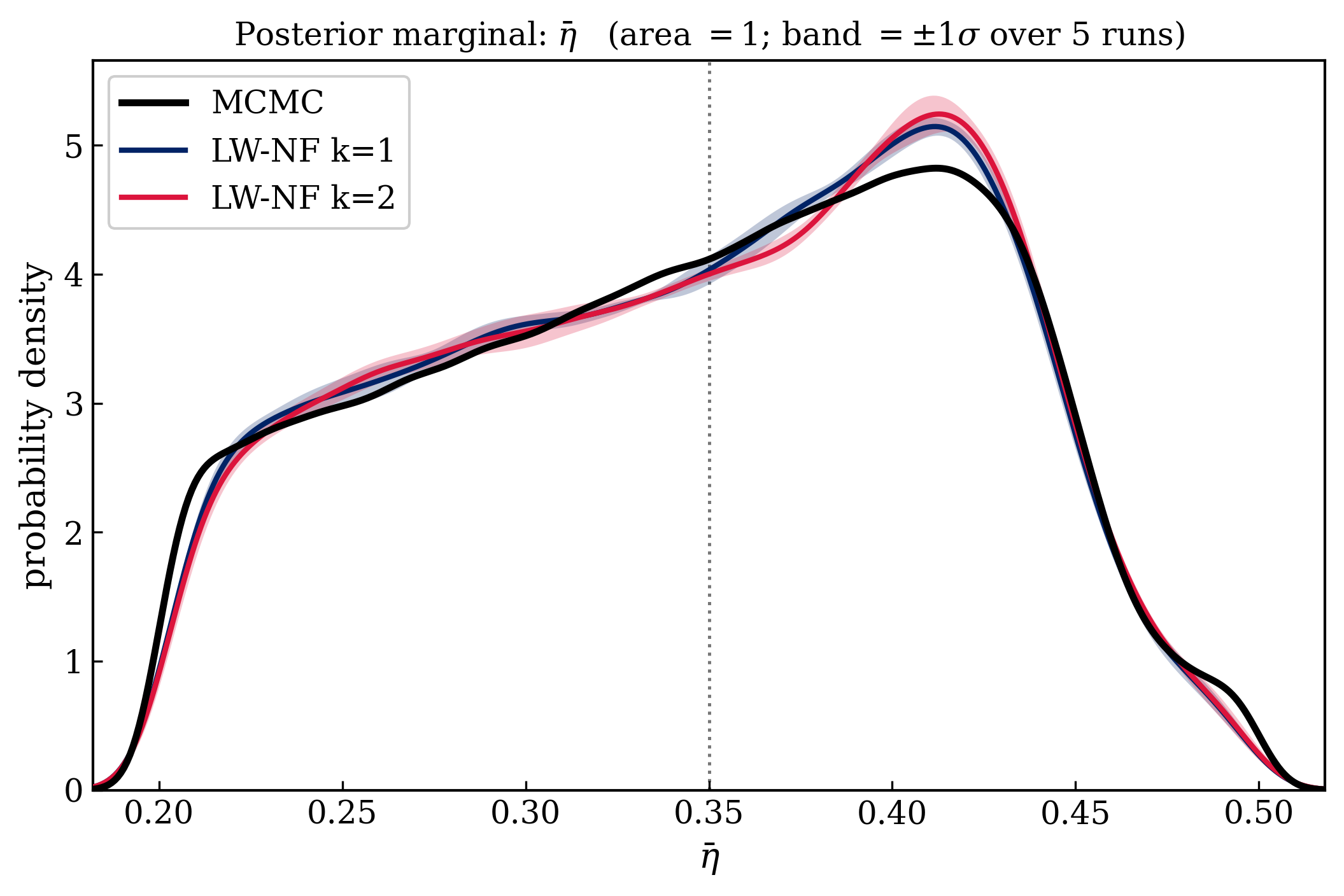}
        \caption{\footnotesize
        Marginal posterior in $\bar\eta$, the soft direction along each
        wedge. The broad, non-Gaussian shape, which rises linearly and
        drops where the first solution leaves the prior near
        $\bar\eta\approx0.45$, is reproduced by both flows within the
        run-to-run band, confirming that the LW-NF captures the elongated
        shape of each mode and not only the discrete $\bar\rho$ structure.}
        \label{fig:marg_eta}
    \end{subfigure}

    \caption{\footnotesize
    One-dimensional marginal posteriors for $\bar\rho$ and $\bar\eta$ comparing
    MCMC and likelihood-weighted normalizing flows.}
    \label{fig:marginals}
\end{figure*}
}

\paragraph{Both bases reproduce the asymmetric posterior}
The exact posterior gives $f_1=0.811$ and the MCMC reference $f_1=0.809$. The flows reproduce this imbalance closely: $f_1=0.824\pm0.001$ for $k{=}1$ and $f_1=0.819\pm0.006$ for
$k{=}2$. Both recover \emph{both} modes at very nearly the correct weight, with mean marginal Jensen--Shannon divergences of $\sim\!10^{-3}$ and mean Wasserstein distances of $\sim\!7\times10^{-3}$ against the reference, i.e., the LW-NF posterior is, to plotting accuracy, the MCMC posterior. The narrow $|V_{cb}|$ peak and the unimodal $g$ marginal are recovered to within the per-dimension transport resolution, confirming that the flows also capture the broken product degeneracy and the soft $\bar\eta$ direction, not merely the two $\bar\rho$ peaks.

\paragraph{The flows carry a small bias that the ensemble spread does not reveal}
The exact value lets us see something that the MCMC reference alone could not. Both bases place slightly too much mass in the dominant mode, by $0.013$ for the Gaussian base and by $0.008$ for the GMM-2 base, that is, by $1$--$1.6\%$ of $f_1$. For the Gaussian base, this is more than ten times the spread of the five flows. The spread of an ensemble trained on the same weighted sample is therefore not a reliable error bar on the mode fraction: all the flows see the same $N_{\rm eff}\approx10^4$ effective points, and they can share the same bias. We show how to remove this bias in \cref{sec:bmeson:is}.

\paragraph{$k{=}2$ is slightly better than $k{=}1$}
Although the two bases are close, the GMM-2 base is the more accurate of the two on almost every measure. Its mode fraction is nearer the reference ($0.819$ vs.\ $0.824$ against $0.809$), its mean JS ($0.0012$ vs.\ $0.0013$)
and mean $W_1$ ($0.0067$ vs.\ $0.0075$) are lower, and its minority-mode centroid is closer to the reference ($d_{\rm Mah}^{(2)}=0.108$ vs.\ $0.117$). The Gaussian base is closer on the dominant-mode Mahalanobis distance to the reference ($d_{\rm Mah}^{(1)}=0.050$ vs.\ $0.053$). Most of these differences lie inside the run-to-run scatter, so the advantage of $k{=}2$ is consistent but modest. We compare the truth-based distances with the exact values, $0.124$ and $0.364$ (\cref{sec:bmeson:exact}), rather than with zero. The MCMC reference reproduces both ($0.128$ and $0.361$), and so do both flows within their spread. The larger truth-based distance of the minority mode is therefore a property of the posterior, not a failure of any method.

\paragraph{A single-Gaussian base is already adequate in this problem}
This near-parity is itself a physically meaningful result. Because the posterior weight is so unequal ($\approx81/19$), the minority mode is a small secondary lobe sitting on top of a dominant one rather than a co-equal peak. Carving such a configuration out of a single Gaussian base is plausibly easier than splitting mass evenly between two balanced, equidistant modes: the $k{=}1$
flow need only grow a modest tail toward $\bar\rho_2$, and the coupling layers are expressive enough to do so once training is stable. The toy problems of \cref{sec:Multimodality} suggest that a mixture base matters more when the modes carry comparable mass and are far apart, but we have not tested this in the physics problem. For the strongly asymmetric 4D posterior, the second component of the base buys only a small, incremental improvement.

\begin{table}[htb]
\centering
\footnotesize
\begin{tabular}{@{}lcccc@{}}
\toprule
Method
& $f_1$
& $f_2$
& $\overline{\rm JS}\downarrow$
& $\overline{W}_1\downarrow$ \\
\midrule
Exact (quadrature)
& $0.811$
& $0.189$
& $-$
& $-$
\\

MCMC (ref.)
& $0.809$
& $0.191$
& $-$
& $-$
\\

LW-NF $k=1$
& $0.824\pm0.001$
& $0.176\pm0.001$
& $0.0013\pm0.0002$
& $0.0075\pm0.0013$
\\

LW-NF $k=2$
& $\mathbf{0.819}\pm0.006$
& $\mathbf{0.181}\pm0.006$
& $\mathbf{0.0012}\pm0.0003$
& $\mathbf{0.0067}\pm0.0021$
\\
\bottomrule
\end{tabular}
\caption{\footnotesize
Mode fractions, with the modes separated at $\bar\rho=0.5$, and mean marginal Jensen--Shannon ($\overline{\rm JS}$) and Wasserstein-1 ($\overline{W}_1$) distances to the well-converged MCMC reference. The exact values are computed by quadrature (\cref{sec:bmeson:exact}). Reference-relative metrics are not given for the references themselves ($-$).
}
\label{tab:posterior_metrics}
\end{table}

\begin{table}[htb]
\centering
\footnotesize
\begin{tabular}{@{}lcccc@{}}
\toprule
Method
& $d_{\rm Mah}^{(1)}\downarrow$
& $d_{\rm Mah}^{(2)}\downarrow$
& $d_{\rm TrMah}^{(1)}\downarrow$
& $d_{\rm TrMah}^{(2)}\downarrow$
\\
\midrule

Exact (quadrature)
& $-$
& $-$
& $0.124$
& $0.364$
\\

MCMC (ref.)
& $-$
& $-$
& $0.128$
& $0.361$
\\

LW-NF $k=1$
& $\mathbf{0.050}\pm0.009$
& $0.117\pm0.030$
& $0.135\pm0.015$
& $0.381\pm0.054$
\\

LW-NF $k=2$
& $0.053\pm0.022$
& $\mathbf{0.108}\pm0.013$
& $\mathbf{0.124}\pm0.014$
& $\mathbf{0.341}\pm0.038$
\\

\bottomrule
\end{tabular}
\caption{\footnotesize
Mode-location metrics. $d_{\rm Mah}^{(i)}$ measures the displacement of the reconstructed mode centroid from the MCMC reference centroid in units of the reference per-mode covariance, while $d_{\rm TrMah}^{(i)}$ measures the displacement from the injected parameter values. The truth-based distance of the exact posterior is not zero, because the prior displaces the mean of each mode; it is the value that a faithful method should reproduce.
}
\label{tab:mode_metrics}
\end{table}

\subsection{Importance-Sampling Correction and the Evidence}
\label{sec:bmeson:is}

The flows are close to the exact posterior, but not equal to it, and in a problem without an exact reference their error would not be visible. However, a flow is not only a sampler. It is also a normalised density that can be evaluated exactly, \cref{eq:logq_jac}. We can therefore remove its error by importance sampling with the flow as the proposal. To keep the weights bounded where the flow has too little mass, we use a defensive mixture of the flow and the prior~\cite{hesterberg1995},
\begin{equation}
    r(\theta) = \alpha\,p(\theta) + (1-\alpha)\,q_\phi(\theta),
    \qquad
    w(\theta) = \frac{p(D|\theta)\,p(\theta)}{r(\theta)} .
    \label{eq:bmeson_isweights}
\end{equation}
Since $r\geq\alpha\,p(\theta)$, every weight is bounded by $p(D|\theta)/\alpha$, which is at most $1/\alpha$ since $\ell(\theta) = -\chi^2/2 \leq 0$. With $N$ draws from $r$, posterior expectations are weighted averages, $\mathbb{E}_{p(\theta|D)}[h] \approx \sum_i w_i h(\theta_i)/\sum_i w_i$, and the evidence is estimated by $\hat Z = \frac{1}{N}\sum_i w_i$. The statistical uncertainties follow from the effective sample size: $\sigma_f \approx \sqrt{f(1-f)/N_{\rm eff}}$ for a fraction $f$, and $\sigma_{\ln Z} \approx \sqrt{1/N_{\rm eff}-1/N}$ for the evidence. This is NIS~\cite{muller2019neural}, and it is the same correction that Dax \textit{et al.} apply to neural posterior estimates in gravitational-wave inference~\cite{dax2023}. In this language, LW-NF is the first step of NIS, in which the proposal is the prior and the weights are the likelihood.

\begin{figure*}[htb]
    \centering
    \includegraphics[width=\textwidth]{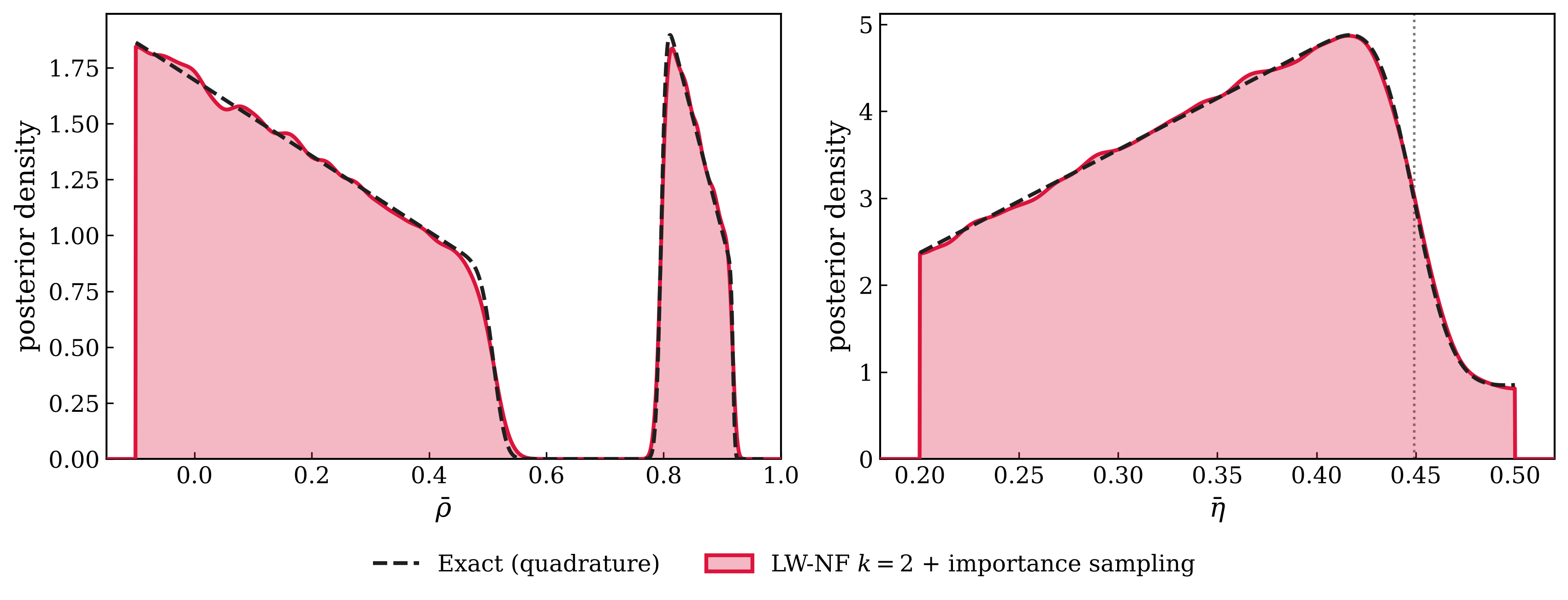}
    \caption{\footnotesize
    Marginal posteriors of $\bar\rho$ (left) and $\bar\eta$ (right) from a $k=2$ flow after the importance-sampling correction of \cref{eq:bmeson_isweights} (red), compared with the exact marginals computed by quadrature (black, dashed). The red curves are Gaussian kernel density estimates of the corrected samples, reflected at the edges of the prior. The dotted line on the right marks $\bar\eta\approx0.449$, where the first solution reaches the lower edge $\bar\rho=-0.10$ of the prior.}
    \label{fig:is_marginals}
\end{figure*}

We apply the correction to a single flow with a GMM-2 base, 12 coupling blocks, hidden layers of width 64, and $c=3$, trained with LW-NF as in \cref{sec:methods:nf}. We then refined the flow by ten rounds of NIS, each with $10^5$ draws from $r$ and $\alpha=0.1$. The efficiency $N_{\rm eff}/N$ of the proposal stayed at about $62\%$ throughout, so the refinement did not improve on the LW-NF flow. For the final estimate, we use $2\times10^5$ draws from $r$, of which $192\,581$ lie inside the prior box. We list the results in \cref{tab:bmeson_is} and compare the marginals with the exact ones in \cref{fig:is_marginals}.

\begin{table}[htb]
\centering
\footnotesize
\begin{tabular}{@{}lcc@{}}
\toprule
Quantity & Exact (quadrature) & LW-NF $k=2$ with IS \\
\midrule
$f_1$, split at $\bar\rho=0.5$ & $0.8111$ & $0.8100\pm0.0011$ \\
$f_1$, split at $\beta=\pi/4$  & $0.8220$ & $0.8210\pm0.0011$ \\
$\ln Z$                         & $-9.4989$ & $-9.5032\pm0.0019$ \\
$N_{\rm eff}$ / draws inside the prior & $-$ & $118\,672/192\,581=0.62$ \\
\bottomrule
\end{tabular}
\caption{\footnotesize
Mode fractions by the two definitions of \cref{sec:bmeson:expect}, and the logarithm of the evidence, from the exact quadrature and from a single $k=2$ flow used as the proposal of \cref{eq:bmeson_isweights} with $\alpha=0.1$. The evidence is given up to the normalisation constant of the Gaussian likelihood, which is the same in both columns. The uncertainties are statistical and follow from $N_{\rm eff}$.}
\label{tab:bmeson_is}
\end{table}

The corrected mode fractions agree with the exact values within their statistical uncertainty, by both definitions, and the bias of $0.008$--$0.013$ in $f_1$ of the uncorrected flows is gone. The evidence agrees with the exact value to $4\times10^{-3}$ in $\ln Z$, that is, to $0.4\%$ in $Z$. This is about twice the statistical uncertainty estimated from $N_{\rm eff}$, so we take that estimate of the uncertainty as approximate; the difference is still small for any practical use of the evidence, for example in a Bayes factor. In \cref{fig:is_marginals}, the corrected marginals lie on the exact curves, including the drop of the $\bar\eta$ density where the first solution leaves the prior.

The real difference between the two steps lies in their cost. The training set of the LW-NF flow needed about $4.5\times10^7$ likelihood evaluations for $N_{\rm eff}\approx9.5\times10^3$, which is about $4700$ evaluations per effective sample. With the trained flow as the proposal, an effective sample costs about $1.6$ evaluations. The first number is fixed by the problem through \cref{eq:ess_chi2}, since it measures how much of the prior volume the posterior occupies. The second is fixed by the quality of the flow. In this problem the likelihood is cheap, and neither number matters in practice. In a problem in which each evaluation runs a spectrum generator, the ratio between them decides whether the method can be used at all. That the adaptive rounds of NIS did not improve the proposal here means that, in four dimensions, the LW-NF flow was already as good a proposal as the refinement could make it with this architecture. We have not yet tested whether adaptation helps in higher dimensions, where the prior is a much poorer proposal and the training set of LW-NF is correspondingly smaller.

\subsection{Discussion}
\label{sec:discussion}

The most important interpretive point is that the $\approx81/19$ mode split is the \emph{correct} answer for this problem, and that its value can be derived. It follows from placing a flat prior on $\bar\rho$ and $\bar\eta$ rather than on $\beta$, which makes the first mode six times wider in every slice of $\bar\eta$ (\cref{eq:slopes}), and from the boundary of that prior, which removes part of the first mode (\cref{eq:weight_ratio}). A method that returned $50/50$ would be wrong. The exact posterior, the MCMC reference, and both flows all find a split close to $81/19$ by the midpoint definition, with the flows overestimating the dominant mode by about one percentage point. In this problem the likelihood is the same on both solutions, and the ratio of their masses comes entirely from the Jacobian between $\bar\rho$ and $\beta$ and from where the prior box ends, so we have to state and examine the choice of prior.

A trained flow is better used as a proposal than as the final answer. Its samples are cheap, but their error is not known without a reference, and, as the Gaussian base shows, the spread of an ensemble of flows can underestimate it. With the importance-sampling correction of \cref{sec:bmeson:is}, the same flow gives the posterior and the evidence with a known statistical uncertainty, at less than two likelihood evaluations per effective sample. We also note what the flow represents. A trained LW-NF represents the posterior for one set of measurements, one prior, and one model. New samples from it cost a single forward pass, but a change of any of these needs new likelihood evaluations and a new training set. The method is therefore not amortised in the sense of simulation-based inference, where one network serves many data sets. A flow trained for one data set could still serve as a proposal for a nearby one, with the weights of \cref{eq:bmeson_isweights} computed with the new likelihood, but we have not tested this.

For the 4D posterior the Gaussian and GMM-2 bases are nearly interchangeable, with $k{=}2$ holding a small, consistent edge on minority-mode accuracy. This should not be read as a general claim that the base is irrelevant. The near-parity here is plausibly a feature of the strong weight asymmetry, which turns the minority mode into a perturbation that a single base can absorb. We expect the mixture base to matter more as the weight balance approaches even or as the inter-mode separation grows, as in the toy problems of \cref{sec:Multimodality}, but this remains to be tested in a physics problem. The 4D problem sits in the regime where that insurance is cheap but not yet essential.

Since both flows recover both modes, the marginal JS and $W_1$ are here faithful summaries of agreement with the reference. The truth-based distance $d_{\rm TrMah}$ (\cref{app:mahal_truth}) is not an absolute measure of accuracy in this problem: the exact posterior itself sits at $d_{\rm TrMah}^{(1)}=0.124$ and $d_{\rm TrMah}^{(2)}=0.364$ from the injected values, because the prior displaces the mean of each mode. The minority mode is nonetheless placed less precisely relative to the reference ($d_{\rm Mah}^{(2)}\approx0.11$ against $0.05$), and its spread over the ensemble is larger, as expected for a mode that carries less than a fifth of the weight, and thus of the effective training set.

\section{Conclusion}
\label{sec:conclusion}

In this work, we showed that minimising the forward KL divergence from the true posterior to the modelled distribution is mathematically equivalent to maximising the model density weighted by the data likelihood. Consequently, a Normalizing Flow trained with likelihood weights approximates the posterior without requiring posterior samples. This objective is not new: it is likelihood weighting followed by weighted maximum-likelihood fitting, and the special case of neural importance sampling with the prior as the proposal. We validated it on 2-dimensional and 3-dimensional benchmark problems. Quantitative evaluations using both KL-divergence and Wasserstein distances confirm that the modelled distributions achieve significant overlap with the true posteriors.

Furthermore, we investigated the effect of the base distribution by replacing the standard Gaussian with a Gaussian Mixture Model (GMM). With a unimodal base, the flow leaves spurious ``bridges'' of probability between well-separated modes, because a flow of finite depth can stretch its base only so far and the forward divergence penalises the misplaced mass only weakly. In our tests, a base with as many components as there are modes gave the lowest distance measures and the highest fidelity reconstruction, while a base with too few components could make the Wasserstein distances worse. Further discussions about topology-aware generative algorithms can be found in~\cite{ngairangbam2025enhancinganomalydetectiontopologyaware, 2019arXiv190903334J}.

We then put the method to work on a curated physics problem: the four-dimensional $B^0\to J/\psi\,K^0$ CP-violation posterior in the Wolfenstein parameters $(\bar\rho,\bar\eta,|V_{cb}|,g)$, a target that is simultaneously multimodal, non-Gaussian, and asymmetric in its mode weights. Its posterior factorises and can be computed exactly, and the $81/19$ split between the modes follows from a short argument: the first mode is $1/\tan^2\beta_1\approx6$ times wider in every slice of $\bar\eta$, and the boundary of the prior reduces this ratio to about $4.6$. Compared with the exact value ($f_1=0.811$) and a long, well-converged MCMC reference ($f_1=0.809$), the flows with the two bases recover both modes at very nearly the correct weight ($f_1=0.824\pm0.001$ for the Gaussian base and $0.819\pm0.006$ for the GMM-2 base), with marginal Jensen--Shannon and Wasserstein distances of order $10^{-3}$. The two-component base is marginally more accurate on most metrics, but the single Gaussian base is already adequate in this strongly asymmetric problem. The remaining bias of $1$--$1.6\%$ in $f_1$ is larger than the spread of the ensemble suggests. Used as a proposal for importance sampling, a single flow removes it and gives the mode fractions and the evidence in agreement with the exact values, at about $1.6$ likelihood evaluations per effective sample, against about $4700$ for its own training set. When projected into observable space, both modes fall onto a single region consistent with the measured CP asymmetry and $|V_{cb}|$, which confirms that the inferred posteriors reproduce the data but does not test the weights of the modes.

Estimating complex parameter spaces is a foundational challenge in High Energy Physics (HEP), prompting the continuous evolution of advanced inference algorithms \cite{Baruah:2025nby,Baruah:2024gwy}. Recently, physicists have increasingly turned to Simulation-Based Inference (SBI) for scenarios where the likelihood is analytically impossible or computationally prohibitive to evaluate. Where the likelihood can be evaluated, likelihood-weighted training of Normalizing Flows offers a simple alternative: the likelihood evaluations are made once, in parallel, and independently of the flow, and the trained flow then serves as a proposal that makes further posterior samples, and the evidence, cheap and checkable.

However, training flows with multi-modal base distributions presents its own challenges, particularly in higher dimensions. The network lacks explicit guidance on which base mode should map to which target mode, leading to combinatorial ambiguity and optimization instability. Thus, care must be taken to initialize or regularize these models effectively.

Our empirical finding, that the best results came when the base matched the number of modes, suggests a promising avenue for future research: adaptive methods that characterise and match the number of modes of an unknown posterior. A second open question is whether the adaptive rounds of NIS, which did not improve the proposal in the four-dimensional problem, help in higher dimensions, where the prior is a much poorer proposal.

\acknowledgments

The author would like to acknowledge support from ANRF (erstwhile DST-SERB), India (grant order no. CRG/2022/003208). The author thanks Sunando Kumar Patra, Bangabasi Evening College, and Subhadeep Mondal, Bennett University, for their helpful insights and guidance on the analysis of distribution topology and reconstruction fidelity, and in preparing this manuscript.

\section*{Code and data availability}
 
The complete implementation, including the likelihood-weighted training pipeline, the 2D and 3D multimodal benchmarks, the 4D $B^0\to J/\psi\,K^0$ analysis, and every script that generates the figures and tables in this paper, is publicly available at\\
\href{https://github.com/rajneilbaruah/lwnf-multimodal-posteriors}{\texttt{github.com/rajneilbaruah/lwnf-multimodal-posteriors}}\\and archived at Zenodo, DOI \href{https://doi.org/10.5281/zenodo.21765049}{10.5281/zenodo.21765049}. The
datasets --- the likelihood-weighted training sets, the reference posterior
samples for every benchmark, the Markov Chain Monte Carlo reference chain, and
the trained flow weights together with the posterior samples and per-run
metadata (random seed, architecture, validation loss and metrics) for all twenty
runs of the 4D sweep --- are archived at DOI
\href{https://doi.org/10.5281/zenodo.21763965}{10.5281/zenodo.21763965}. The
repository documents which script produces each figure and table, and records
the hardware and wall-clock time of every training run reported here.

 
\appendix
 
\section{Statistical Metrics}
\label{app:metrics}
 
All metrics are computed with respect to the MCMC posterior as the reference
distribution unless otherwise noted. We write $p$ for the MCMC reference and
$q$ for the method under evaluation.

\subsection{Mode Detection and Weight Fractions}
\label{app:mode}
A mode is declared \emph{found} if at least one sample from a given
method lies within $0.30\times\|\mu_1-\mu_2\|$ of the true mode
location, where $\|\mu_1-\mu_2\|$ is the inter-mode Euclidean distance.
The weight fractions $f_1$ and $f_2$ are estimated by splitting samples
at the midpoint of the dimension with the largest inter-mode separation.

\subsection{Jensen-Shannon Divergence}
\label{app:js}
The Jensen--Shannon divergence~\cite{Lin:1991} is a symmetrised, bounded
version of the Kullback--Leibler divergence,
\begin{equation}
    {\rm JS}(p\|q)
    = \frac{1}{2}\,D_{\rm KL}(p\|m)
    + \frac{1}{2}\,D_{\rm KL}(q\|m),
    \qquad
    m=\tfrac{1}{2}(p+q),
    \label{eq:js}
\end{equation}
where $D_{\rm KL}$ is the Kullback-Leibler divergence~\cite{Kullback:1951}.
${\rm JS}\in[0,1]$, with $0$ for identical distributions.
We compute JS marginally using 80-bin histograms with regulariser
$\epsilon=10^{-10}$ and report the mean $\overline{\rm JS}$.

\subsection{Wasserstein-1 Distance}
\label{app:w1}
The Wasserstein-1 (earth-mover's) distance~\cite{Villani:2008, Peyre:2019} is
\begin{equation}
    W_1(p,q)
    = \inf_{\gamma\in\Pi(p,q)}
    \mathbb{E}_{(x,y)\sim\gamma}\!\left[\|x-y\|\right].
    \label{eq:w1}
\end{equation}
For one-dimensional distributions $W_1$ equals the $L^1$ distance
between CDFs. We compute $W_1$ marginally and report $\overline{W}_1$.

\subsection{Sliced Wasserstein Distance}
\label{app:sw}
The sliced Wasserstein distance~\cite{Rabin:2011, Bonneel:2015} replaces the costly $d$-dimensional transport problem of \cref{eq:w1} by an average of one-dimensional $W_1$ distances along random projections,
\begin{equation}
    \mathrm{SW}_1(p,q)
    = \int_{\mathbb{S}^{d-1}}
      W_1\!\left(p_\omega,\,q_\omega\right)\,
      \mathrm{d}\sigma(\omega),
    \label{eq:sw}
\end{equation}
where $p_\omega$ is the distribution of the projection $\langle\omega,x\rangle$ for $x\sim p$, and $\sigma$ is the uniform measure on the unit sphere $\mathbb{S}^{d-1}$. Each projected $W_1$ reduces to the $L^1$ distance between one-dimensional CDFs. We approximate the integral by Monte Carlo over random directions and report the mean $\overline{\mathrm{SW}}_1$. Because it aggregates many directions rather than the coordinate axes alone, $\overline{\mathrm{SW}}_1$ is sensitive to inter-dimensional structure that the marginal $\overline{W}_1$ misses, while remaining far cheaper than the full Wasserstein distance.

\subsection{MCMC-Referenced Mahalanobis Distance}
\label{app:mahal}
The Mahalanobis distance~\cite{Mahalanobis:1936} measures the displacement
of a reconstructed mode centroid from the reference centroid in units of the reference covariance,
\begin{equation}
    d_{\rm Mah}^{(i)}
    = \sqrt{
        \bigl(\hat\mu_q^{(i)}-\hat\mu_p^{(i)}\bigr)^\top
        \!\!\bigl(\hat\Sigma_p^{(i)}\bigr)^{-1}
        \!\!\bigl(\hat\mu_q^{(i)}-\hat\mu_p^{(i)}\bigr)
    },
    \label{eq:mahal}
\end{equation}
where $\hat\mu^{(i)}$ and $\hat\Sigma^{(i)}$ are the empirical mean
and covariance of samples assigned to mode $i$, and $p$ denotes the
MCMC reference.
A value of $d_{\rm Mah}=0$ indicates that the method centroid coincides
exactly with the MCMC centroid; $d_{\rm Mah}=1$ indicates a displacement
of one MCMC posterior standard deviation. This metric is undefined for the MCMC reference itself (zero by construction) and measures agreement with the reference rather than the truth.

\subsection{Truth-Based Mahalanobis Distance}
\label{app:mahal_truth}
To assess accuracy against known ground truth — applicable because
the true mode locations are specified by the benchmark construction —
we define
\begin{equation}
    d_{\rm TrMah}^{(i)}
    = \sqrt{
        \bigl(\hat\mu_q^{(i)}-\theta^{{\rm mode},(i)}\bigr)^\top
        \!\!\bigl(\hat\Sigma_p^{(i)}\bigr)^{-1}
        \!\!\bigl(\hat\mu_q^{(i)}-\theta^{{\rm mode},(i)}\bigr)
    },
    \label{eq:mahal_truth}
\end{equation}
where $\theta^{{\rm mode},(i)}$ is the true parameter vector at mode $i$
(Table~\ref{tab:mode_metrics}) and $\hat\Sigma_p^{(i)}$ is the MCMC per-mode covariance used as normaliser. Unlike the MCMC-referenced Mahalanobis, $d_{\rm TrMah}$ is defined for the MCMC reference itself. It is not zero even for the exact posterior, since the prior displaces the mean of each mode from the injected point (\cref{sec:bmeson:exact}); a faithful method reproduces the exact value, $0.124$ and $0.364$, not zero. Per-mode sample assignment uses the same midpoint splitting as the weight fraction calculation.

\section*{Declaration}
During the preparation of this work, the author used Anthropic's Sonnet and Opus for grammatical proofreading, sentence restructuring, and building the GitHub/Zenodo Repository. The author reviewed and edited the output as needed and take full responsibility for the content of the published article.

\bibliographystyle{JHEP}
\bibliography{biblio.bib}

\end{document}